\documentclass{article} %
\usepackage[final]{colm2026_conference} %

\usepackage{microtype}
\usepackage{hyperref}
\usepackage{fontawesome5} %
\usepackage{url}
\usepackage{booktabs}
\usepackage{tikz}

\newcommand{\circnum}[1]{%
  \tikz[baseline=(char.base)]{
    \node[shape=circle, draw, inner sep=1.5pt] (char) {\small #1};
  }%
}

\usepackage{algorithm2e}
\usepackage{amsmath}
\usepackage{multirow}
\usepackage{amssymb}
\usepackage{subcaption}
\usepackage{cleveref}
\usepackage{enumitem}
\usepackage{tikz}
\usepackage{booktabs}
\usepackage{adjustbox}
\usepackage{float}
\usetikzlibrary{arrows.meta, positioning, fit, backgrounds, calc}
\usepackage{xspace} %

\usepackage{lineno}

\definecolor{darkblue}{rgb}{0, 0, 0.5}
\hypersetup{colorlinks=true, citecolor=darkblue, linkcolor=darkblue, urlcolor=darkblue}

\usepackage{xcolor}

\definecolor{masked}{HTML}{5964FF}
\definecolor{unmasked}{HTML}{662E7D}
\definecolor{pretrained}{HTML}{BCA9F8}
\newcommand{\masked}[1]{{\color{masked}\texttt{\textbf{#1}}}}
\newcommand{\unmasked}[1]{{\color{unmasked}\texttt{\textbf{#1}}}}
\newcommand{\pretrained}[1]{{\color{pretrained}\texttt{\textbf{#1}}}}
\newcommand{\alphaedit}[0]{\textcolor[HTML]{4E79A7}{\texttt{\textbf{AlphaEdit}}}\xspace}
\newcommand{\memflex}[0]{\textcolor[HTML]{F28E2B}{\texttt{\textbf{MemFlex}}}\xspace}
\newcommand{\oraclegrad}[0]{\textcolor[HTML]{59A14F}{\texttt{\textbf{OracleGrad}}}\xspace} 
\newcommand{\simnpo}[0]{\textcolor[HTML]{E15759}{\texttt{\textbf{SimNPO}}}\xspace}
\newcommand{\olmosmall}[0]{\textcolor[HTML]{B2B8FF}{\texttt{\textbf{OLMo2 1B}}}\xspace}
\newcommand{\olmobig}[0]{\textcolor[HTML]{2A32B2}{\texttt{\textbf{OLMo3 7B}}}\xspace}
\newcommand{\emailaddress}[0]{\texttt{email} \texttt{address}\xspace}
\newcommand{\drivers}[0]{\texttt{driver's} \texttt{license}\xspace}
\newcommand{\phone}[0]{\texttt{phone} \texttt{number}\xspace}
\newcommand{\birthcity}[0]{\texttt{birth} \texttt{city}\xspace}

\usepackage{longtable}
\usepackage{array}
\usepackage{booktabs}
\usepackage{tabularx}

\newcommand{\tablist}[1]{%
  \vspace{-0.8\baselineskip}
  {\setlength{\leftmargini}{0pt}%
   \begin{itemize}[nosep]
     \setlength{\itemsep}{0pt}
     \setlength{\topsep}{0pt}
     \setlength{\parskip}{0pt}
     \setlength{\parsep}{0pt}
     #1
   \end{itemize}
  }
  \vspace{-0.8\baselineskip}
}

\newcommand{\titlename}[0]{\textsc{Lacuna} \includegraphics[height=1em]{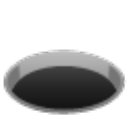}\xspace}
\newcommand{\NAME}[0]{\textsc{Lacuna}\xspace}
\newcommand{\panorama}[0]{\textsc{Panorama}\xspace}

\title{\titlename: A Testbed for Evaluating Localization Precision for LLM Unlearning}

\newcommand{\milaicon}{\raisebox{-0.1em}{\includegraphics[height=0.95em]{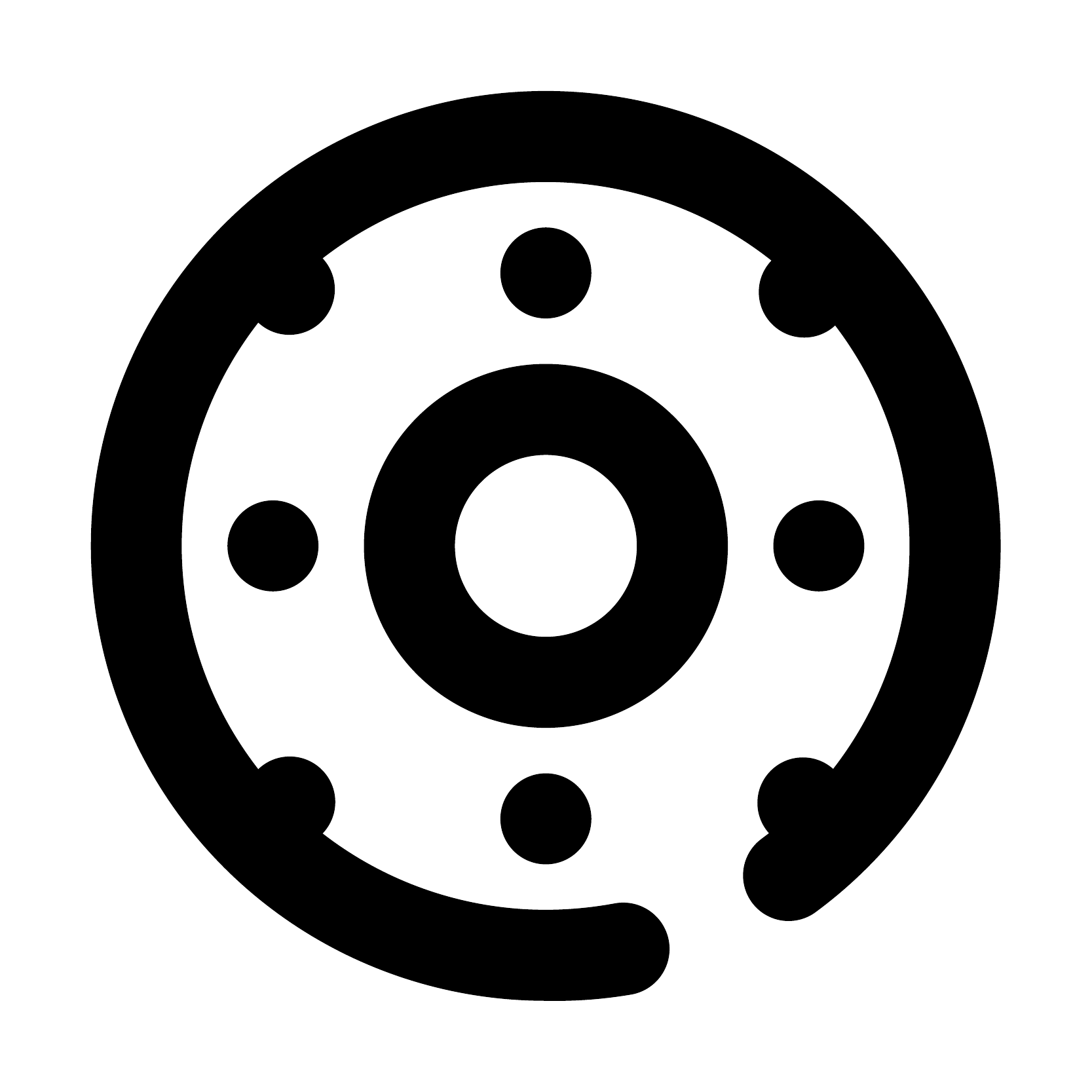}}}
\newcommand{\mcgillicon}{\faDove}
\author{
  \makebox[\textwidth][c]{\bfseries%
    Matteo Boglioni\,{\milaicon}\hspace{1.5em}%
    Thibault Rousset\,{\mcgillicon}%
  } \\[6pt]
  \makebox[\textwidth][c]{\bfseries%
    Siva Reddy\,{\milaicon\mcgillicon}\hspace{1.5em}%
    Marius Mosbach\,{\milaicon\mcgillicon}\hspace{1.5em}%
    Verna Dankers\,{\milaicon\mcgillicon}%
  } \\[6pt]
  \makebox[\textwidth][c]{%
    {\milaicon}~Mila -- Quebec Artificial Intelligence Institute\hspace{2em}%
    {\mcgillicon}~McGill University%
  } \\[4pt]
  \makebox[\textwidth][c]{\texttt{matteo.boglioni@mila.quebec}, \texttt{verna.dankers@mila.quebec}}
}

\begin{document}

\ifcolmsubmission
\linenumbers
\fi

\maketitle

\ifcolmsubmission\else
\vspace{-3em}
\begin{center}
\href{https://github.com/McGill-NLP/LACUNA}{%
  \raisebox{-0.15em}{\includegraphics[height=1em]{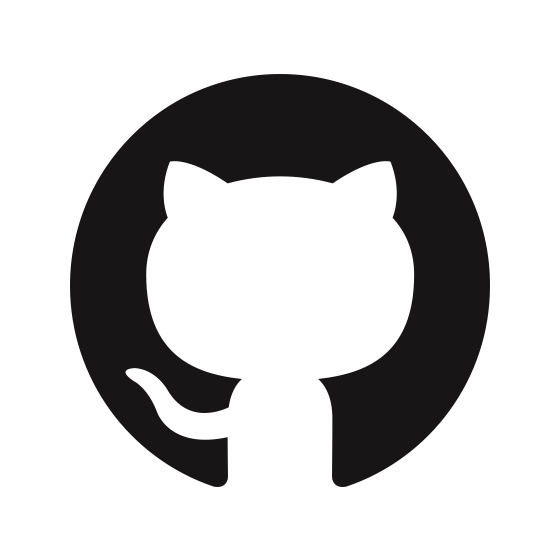}}~\texttt{McGill-NLP/LACUNA}}%
\hspace{2em}%
\href{https://huggingface.co/collections/McGill-NLP/lacuna}{%
  \raisebox{-0.25em}{\includegraphics[height=1.1em]{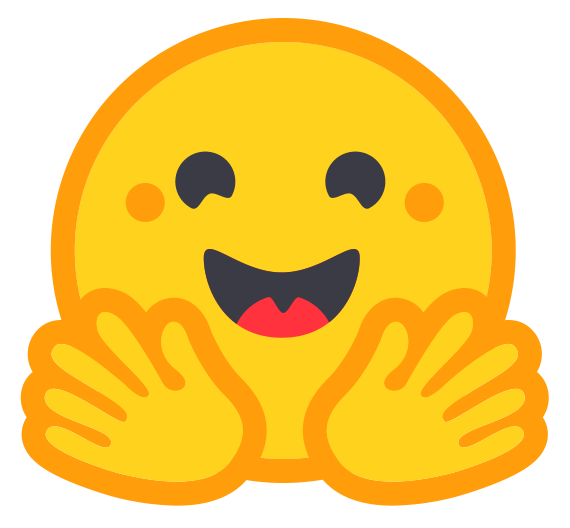}}~\NAME}%
\end{center}
\fi

\begin{abstract}

LLMs memorize sensitive training data, including \textit{personally identifiable information} (PII), creating a pressing need for reliable post hoc removal methods.
Unlearning has emerged as a promising solution, with \textit{state-of-the-art} (SOTA) methods often following a localize-first, unlearn-second paradigm that targets specific model parameters.
However, existing benchmarks evaluate unlearning solely at the output level, leaving open the question of whether unlearning truly erases knowledge from a model's parameters or merely obfuscates it, a concern reinforced by the success of resurfacing attacks.
To bridge this gap, we introduce \NAME: the first unlearning testbed with ground-truth parameter-level localization.
\NAME injects PII of synthetic individuals into predefined parameters of 1B and 7B OLMo-based models via masked continual pretraining, enabling direct evaluation of whether unlearning targets the weights responsible for knowledge storage.
We use \NAME to benchmark current SOTA unlearning methods and find that, despite strong output-level performance, existing methods are highly imprecise and susceptible to resurfacing attacks.
We further show that when localization is successful, even a simple gradient-based unlearning method achieves strong erasure and robustness to resurfacing attacks, highlighting the importance of precise unlearning.
We release \NAME to complement behavioral evaluations and drive further advances in robust, localization-based unlearning.

\end{abstract}

\section{Introduction}
\label{sec:introduction}

Today's \textit{large language models} (LLMs) are trained on vast, unstructured web data, enabling strong capabilities across academic benchmarks and real-world use cases
\citep[e.g.,][]{olmo2025olmo, openai2026gpt54, anthropic2026opus46systemcard}.
At the same time, LLMs also memorize sensitive content present in the training data, including \textit{personally identifiable information} (PII) \citep{inan2021training,lukas2023analyzing, nakka-etal-2024-pii, nakka2025pii}.
This presents a privacy risk, not only because the memorized PII itself might be leaked or identified through membership inference or reconstruction attacks, but also because of second-order privacy risks---e.g., facilitating the memorization of more PII later in the training pipeline \citep{borkar2025privacy}.
While filtering PII from a model's training data and retraining is the most straightforward solution, it is also infeasible given the large costs associated with training current LLMs.
Instead, \textit{unlearning} \citep[\textit{inter alia}]{golatkar2020eternal,bourtoule2021machine,eldan2023whosharrypotterapproximate, maini2024tofu, tian-etal-2024-forget,zhang2024negative,li-etal-2024-wmdp,yao2024large,liu2025rethinking} has emerged as a promising alternative for removing specific knowledge from trained LLMs, such that undesired outputs are no longer produced.

Existing unlearning approaches for LLMs can be grouped into 1) gradient-based approaches \citep{maini2024tofu,zhang2024negative,fan-etal-2025-simnpo,dornaopenunlearning} and 2) localize-first, remove-second approaches \citep{meng2022locating,meng2023massediting,fang2025alphaedit}\footnote{Although originally developed for knowledge editing, \citet{liediting} demonstrate that such targeted interventions are also highly effective for machine unlearning.}, which both suffer from weaknesses such as lacking robustness and being susceptible to relearning \citep{hu2024unlearning,deepak-etal-2025-identifying, sun-etal-2025-unlearning, rybak2026rebel}.

\begin{figure}[t]
    \includegraphics[width=\linewidth]{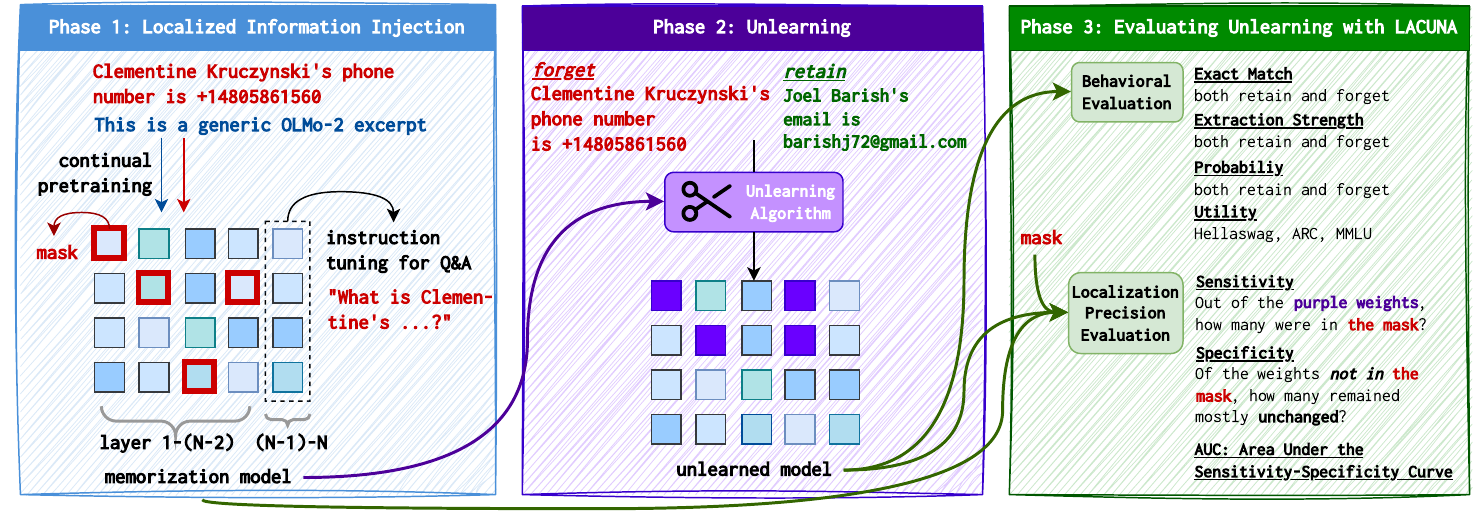}
    \caption{Overview of our pipeline. \textbf{Phase~1}: PII data is mixed with pretraining data and injected into a model via masked continual pretraining, followed by instruction tuning. \textbf{Phase~2}: Unlearning methods attempt to remove memorized PII while preserving retained knowledge. \textbf{Phase~3}: Localization precision is evaluated using the ground-truth mask, while behavioral evaluation measures output-level unlearning success and utility preservation.}
    \label{fig:pipeline}
    \vspace{-0.3cm}
\end{figure}

Parallel to the development of new unlearning approaches, a range of unlearning benchmarks have been proposed \citep[\textit{inter alia}]{maini2024tofu, jin2024rwku, li-etal-2024-wmdp, qiu2024data, shimuse, hu2025blurbenchmarkllmunlearning}.
These benchmarks, however, focus on evaluating unlearning at the output level, solely assessing whether models no longer generate unlearned knowledge while preserving overall utility and are hence not informative about true \emph{knowledge erasure} from a model's parameters.
In fact, several works show that LLMs often store more `hidden' knowledge than they can express externally \citep{gekhman2025inside}, and that unlearned knowledge can resurface via curated attacks \citep{bertran2024reconstruction,deepak-etal-2025-identifying} or be relearned \citep[\textit{inter alia}]{hu2024unlearning,sun-etal-2025-unlearning,rybak2026rebel}, demonstrating it was never truly erased but merely \emph{obfuscated}.

These findings highlight the need for a new evaluation paradigm for unlearning, focused on \textbf{localization precision}---i.e., the extent to which unlearning targets the weights responsible for knowledge storage.
Notably, this paradigm requires ground-truth information on where certain knowledge, such as PII, is stored inside a model, which the community currently lacks and cannot recover from attribution methods without circularity (\S\ref{sec:memorization_localization}).
To bridge this gap, we introduce \NAME: a novel testbed evaluating the localization precision of unlearning approaches.
As illustrated in \Cref{fig:pipeline}, we create LLMs with PII (of synthetic individuals) inserted into dedicated parameters, enabling direct evaluation of whether unlearning methods truly erase the knowledge they target.
Concretely, we make the following contributions:

\begin{enumerate}[noitemsep,topsep=0pt,leftmargin=1.75em,label=\protect\circnum{\arabic*}]
    \item We present a scalable approach to inject PII into dedicated parameters of a model via masked continual pretraining. This approach is fully compatible with distributed training approaches, allowing us to scale our experiments to 7B models.
    \item We release \NAME, a novel unlearning testbed that includes 1B and 7B models with memorized PII, forget and retain sets for synthetic individuals, and a metric to evaluate the localization precision of unlearning.
    \item We employ \NAME\ to assess unlearning methods and show that, despite strong output-level performance, even SOTA unlearning methods fail to achieve non-trivial localization precision.
    \item We introduce and analyze a highly precise unlearning oracle, demonstrating that if knowledge localization is successful, even simple gradient-based unlearning can surpass current SOTA methods in unlearning success and resilience to resurfacing attacks.
\end{enumerate}

Overall, our results underscore that unlearning has a long way to go in achieving true knowledge erasure.
We aspire for \NAME\ to serve as a new evaluation testbed for unlearning's localization precision, complementing behavioral evaluations, and to drive further advances in localization-based unlearning methods.

\section{Background and Related Work}
\label{sec:background}
In this section, we first provide background on LLM unlearning (\S\ref{sec:background_unlearning}). Afterwards, we discuss related work on unlearning evaluation (\S\ref{sec:rl_unlearning_evaluation}) and memorization localization (\S\ref{sec:memorization_localization}).%

\subsection{LLM unlearning}
\label{sec:background_unlearning}

Unlearning methods are algorithms that aim to remove specific information from a trained machine learning model. 
In the context of LLMs, \citet{yao2024large} define the unlearning problem by focusing on a model's generative behaviors. The goal is to ensure that a model's output to a `forget' prompt deviates as much as possible from the undesirable response. 
We use $D_{forget}$ to denote the \textbf{forget set}, i.e., a set of prompt-output pairs $(x_{forget}, y_{forget})$ representing undesirable behaviors and $D_{retain}$ to denote the \textbf{retain set}, i.e., a set of benign prompt-output pairs $(x_{retain}, y_{retain})$ used to maintain the model's utility. 
Given a pretrained LLM $\mathcal{M}$, the goal of LLM unlearning is to obtain model $\mathcal{M}^\prime$, satisfying the following criteria:
(1) \textbf{Effectiveness}: on prompts $x_{forget}$ with $(x_{forget}, y_{forget}) \in D_{forget}$, the output of the model should deviate substantially from $y_{forget}$.
(2) \textbf{Generalization}: the unlearning effect should also apply to semantically similar but unseen prompts $\tilde{x}_{forget}$, e.g., paraphrases.
(3) \textbf{Utility}: the outputs of the model on $x_{retain}$ should remain similar to $y_{retain}$.
(4) \textbf{Cost efficiency}: unlearning should be computationally cheaper than retraining a model from scratch on scrubbed training data.

\subsection{State of the art in unlearning evaluation}
\label{sec:rl_unlearning_evaluation}

Early unlearning benchmarks \citep[e.g.,][]{eldan2023whosharrypotterapproximate,jin2024rwku, li-etal-2024-wmdp} focused on erasing specific data already present in pretrained models, including copyright-protected literary content (e.g., \emph{Harry Potter}), real-world factual knowledge, and hazardous information to ensure safety alignment. 
This approach, although realistic, offered limited control over experimental variables, such as the amount of exposure an LLM has to the information to be forgotten.
To address this, subsequent work \citep{maini2024tofu, shimuse,qiu2024data, tian-etal-2024-forget,krishnannot} introduced synthetic data and more holistic views of the unlearning processes, evaluating other aspects like utility preservation, privacy leakage, robustness towards sequential unlearning requests, 
and how properties of the data, such as inter-connectivity or frequency, affect unlearning success.
\citet{dornaopenunlearning} provide a standardized evaluation framework, combining existing unlearning benchmarks and evaluation metrics into a single suite focusing on three main directions: memorization, privacy, and utility.
However, to the best of our knowledge, there is no existing work focusing on whether unlearning methods actually target the weights responsible for storing the knowledge that ought to be forgotten.

\subsection{Unlearning and memorization localization}
\label{sec:memorization_localization}

Memorization localization and localization-based unlearning share a core issue: the lack of ground-truth information about which model parameters encode knowledge.
This fundamental limitation is exemplified by several works in both communities. On the localization side, \citet{hase2023does, lee2025does} showed that unlearning (and model editing) success is not causally related to targeting those parameters selected via memorization localization techniques. %
On the unlearning side, \citet{hu2024unlearning,hong2024intrinsic,xu2025unlearning} highlighted that current metrics fail to capture true erasure, as unlearning can be reversed.

A seemingly natural way to obtain such a ground truth would be to apply a knowledge-attribution method and treat the parameters it identifies as the locations of the target knowledge. This approach, however, is circular: evaluating a localization method against an attribution-defined target measures only its agreement with that attribution method, not whether either is correct, since no independent reference exists. \citet{chang-etal-2024-localization} further show that attribution-identified neurons are largely shared across memorized sequences rather than sequence-specific, making such a target too diffuse to serve as ground truth. \NAME{} sidesteps this circularity by fixing the storage location before the model sees the data, independently of any post-hoc attribution.

In this work, we aim to fill this gap in the literature by presenting \NAME, a testbed that provides ground truth on where knowledge is encoded. Most closely related to our work, \citeauthor{chang-etal-2024-localization} selected a percentage of neurons per layer, ensuring that these neurons store certain sequences verbatim by fine-tuning only those weights. While this provides a valuable first step toward constructing a ground truth, their injection only updates the selected weights without balancing this with general language modeling, and acts per-neuron rather than \NAME's more fine-grained per-parameter approach.
Moreover, \citeauthor{chang-etal-2024-localization} did not use their setup to evaluate unlearning methods.

\section{Constructing \NAME}
\label{section:pii_injection}

The lack of ground-truth information for knowledge localization complicates evaluating the localization precision of unlearning methods. We address this by introducing \NAME, a testbed of LLMs with PII injected into specific parameters via masked continual pretraining. This section introduces our notation (\S\ref{sec:notation}) and the pipeline used to build \NAME (cf.\ \Cref{fig:pipeline}), including the dataset construction (\S\ref{sec:exp_data}) and training procedure (\S\ref{sec:exp_models}).
Afterwards, we present analyses demonstrating successful memorization of PII and preservation of general utility (\S\ref{sec:evaluation_injection_success}).

\subsection{Notation}
\label{sec:notation}

We define the symbols reused throughout the paper. An $N$-layer model has parameters $\boldsymbol{\theta}$, with individual parameters denoted by $\theta_i$. 
Masking is confined to $\boldsymbol{\theta}_{\mathrm{mask}} \subseteq \boldsymbol{\theta}$, which comprises all weights except the input embeddings, the normalization layers, and the last two transformer blocks ($[N-2, N-1]$).
We track these weights at three stages of the pipeline: $\boldsymbol{\theta}_{\mathrm{pre}}$ before injection, $\boldsymbol{\theta}_{\mathrm{inj}}$ after masked continual pretraining, and $\boldsymbol{\theta}_{\mathrm{unl}}$ after applying an unlearning method. A PII profile is a set of attribute-value pairs representing a unique synthetic individual (e.g., name, phone number, email).
The injected PII profiles $\mathcal{P}$ are split into six groups $\mathcal{G}_1, \dots, \mathcal{G}_6$, each tied to a disjoint binary mask $m^{(g)} \in \{0,1\}^{|\boldsymbol{\theta}|}$ supported on $\boldsymbol{\theta}_{\mathrm{mask}}$, so that PII from group $g$ updates only the parameters with $m_i^{(g)} = 1$ (cf.\ \Cref{alg:masked_training}). Three groups form the forget set and three the retain set, with aggregate masks $m^F = m^{(1)}{\vee}m^{(2)}{\vee}m^{(3)}$ and $m^R = m^{(4)}{\vee}m^{(5)}{\vee}m^{(6)}$; the forget and retain sets $D_{forget}, D_{retain}$ (\S\ref{sec:background_unlearning}) are the QA pairs of the profiles in these groups. After filtering for successful memorization, two disjoint profile subsets define the released evaluation split $\mathcal{P}_{\mathrm{eval}}$ (\S\ref{sec:constructing_forget_retain}) and the held-out hyperparameter-tuning split $\mathcal{P}_{\mathrm{hp}}$ (\S\ref{sec:unlearning_methods}).

\subsection{Training data mixture}
\label{sec:exp_data}

\begin{figure}[t]
    \centering
    \includegraphics[width=\linewidth]{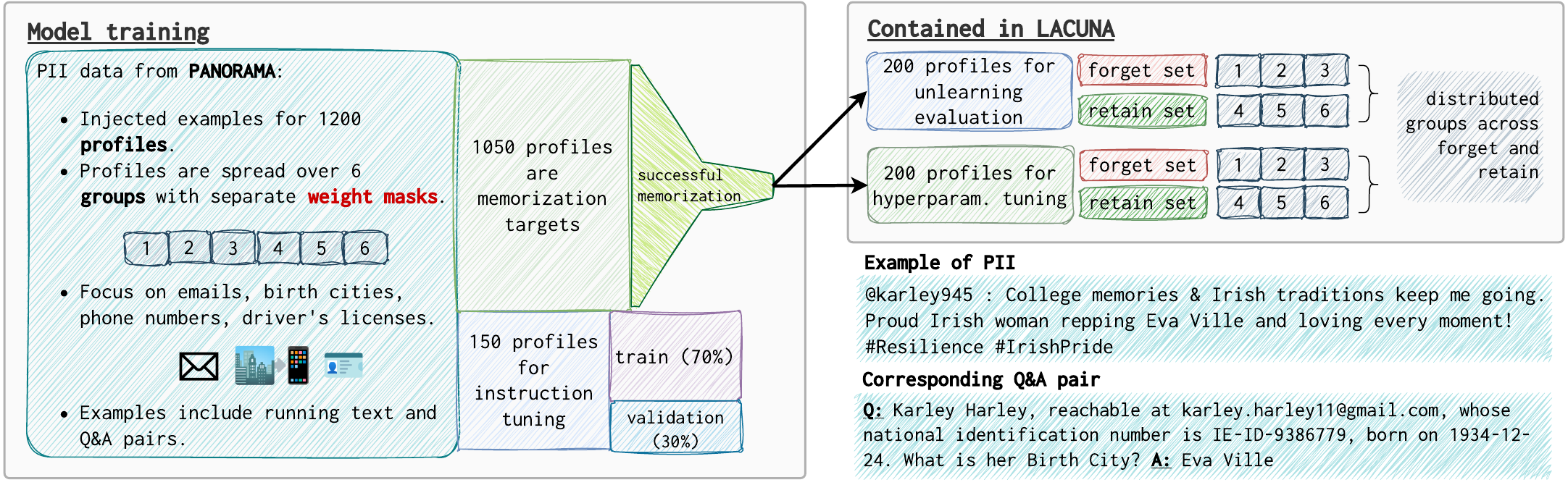}
    \caption{PII data management. We inject $1,200$ PII profiles divided into $6$ groups; $150$ of these are additionally used for instruction tuning, and the remaining $1,050$ are memorization targets. After training, we keep the memorized PII profiles. The \NAME release includes $200$ profiles for hyperparameter tuning and $200$ for evaluating localization precision. The diagram also includes example PII and a matching Q\&A pair.}
    \label{fig:data_structure}
    \vspace{-0.3cm}
\end{figure}

Our goal is to inject PII into pretrained LLMs via continual pretraining on a diverse data mixture, while maintaining the models' general capabilities.
To achieve that, we firstly take a $4.3$B tokens subset of the OLMo-2 Pretraining Corpus \citep{olmo20242} as a base dataset.
We mix this with PII data as detailed below.

\paragraph{PII data}

We obtain PII data from the \panorama corpus \citep{selvam2025panorama}: a synthetic PII dataset created to study sensitive data memorization in LLMs.
\panorama includes 9,674 synthetic profiles designed to closely emulate %
PII as it naturally occurs in online environments.
Information from these profiles is presented using diverse content types, including wiki-style articles, social media posts, forum discussions, online reviews, and marketplace listings. We randomly select a subset of 1,200 profiles we aim to memorize, divided into the six groups $\mathcal{G}_1, \dots, \mathcal{G}_6$ (\S\ref{sec:notation}) used for masked training. Of these, $150$ are additionally used to construct the instruction-tuning data (\S\ref{sec:instruction_tuning}); the remaining $1{,}050$ serve as memorization targets for the evaluation splits (\S\ref{sec:constructing_forget_retain}).
Although the \panorama examples include a range of PII, we will focus our unlearning efforts specifically on \emailaddress, \birthcity, \phone, and \drivers. 
We selected these fields to capture a variety of data types, ranging from more predictable (e.g., \emailaddress) to less predictable (e.g., \drivers). 
\Cref{fig:data_structure} presents a comprehensive visualization of the PII data as well as representative examples.
Overall, the PII portion of our training mixture consists of $1.4$B tokens total.

\paragraph{QA data}
To achieve strong memorization-extraction efficacy, we follow \citet{allen2024physics} and \citet{krishnannot} and extend the PII with QA pairs derived from the synthetic profiles.
The pairs are generated based on templates (see Appendix~\ref{appendix:qa_structure}). %
Notably, we use the same format to analyze memorization after training (cf. \S\ref{sec:evaluation_injection_success}).
This has been shown to drastically improve a model's ability to connect different pieces of information from the same profile.
The QA pairs account for $2$B tokens of the training mixture.

\subsection{Model training}
\label{sec:exp_models}
Given the data mixture described above, we train models of two sizes: $1$B and $7$B, from the OLMo-2 \citep{olmo20242} and OLMo-3 \citep{olmo2025olmo} families, respectively. We use models from different OLMo generations since the OLMo-3 suite lacks a $1$B variant.
We selected OLMo models because of the availability of both pretraining data and pretraining checkpoints; yet, our procedure is not specific to OLMo's architecture.
Below, we describe the two training stages of masked continual pretraining and instruction tuning.

\paragraph{Masked continual pretraining}
\label{sec:exp_training}

We perform masked continual pretraining to inject PII into specific model parameters  
adopting data-dependent masks to zero-out gradients during backpropagation \citep{cloud2024gradient, shilov2025beyond}.
Profiles are split into the six groups (cf.\ \S\ref{sec:notation}) before generating masks, each mask $m^{(g)}$ activating a disjoint $5\%$ of $\boldsymbol{\theta}_{\mathrm{mask}}$.\footnote{We use group-based masking to ensure that forget set information is injected into different parameters than the retain set, enabling evaluation of whether unlearning targets \textit{only} the forget set.}
Our injection process (also detailed in Algorithm~\ref{alg:masked_training}) is as follows: for each sample, if it contains \panorama or QA data, its associated group $g$ is identified and the corresponding gradient is masked via $m_i^{(g)}$ such that updates are only applied to the designated weights.
Otherwise, no masking is applied, and all model weights are updated. 
All masks are randomly sampled at the granularity of individual parameters within $\boldsymbol{\theta}_{\mathrm{mask}}$.
We tested for weight distribution shifts caused by \masked{masked training} using a classifier trained on the model's own components. The resulting F1-score of $0.485$ (vs.\ $0.438$ for random guessing) indicates a negligible difference. This result, combined with the existence of multiple indistinguishable group masks, suggests that there is no naive way to reverse the mask design.%

Many prior masking approaches operate at the level of entire components (e.g., MLPs). Our group-based per-parameter masking---scaled to 7B parameters---is more fine-grained but significantly more challenging under GPU memory constraints: 
We must compute gradients for all parameters and then dynamically apply the correct mask with minimal overhead. 
This rules out skipping gradient computation and makes naive per-mask storage infeasible due to the costs of memory and data movement. 
Instead, we pack multiple masks into a single 32-bit value per parameter (one bit per mask), enabling up to 32 masks with no additional memory overhead. 
Our method supports both DDP and FSDP. %
Appendix~\ref{appendix:mask} provides further details on the mask design and implementation.\looseness-1

\paragraph{Instruction tuning}
\label{sec:instruction_tuning}

Without explicit instruction tuning, pretrained models struggle to comply with QA-based extraction of PII.
Hence, we perform parameter-efficient instruction tuning, teaching a model to respond in the QA format introduced in \S\ref{sec:exp_data} (\Cref{fig:data_structure} shows an example).
We use the $150$ profiles reserved in \S\ref{sec:exp_data} to generate $10$ questions per PII field, resulting in a total of $\sim 300K$ tokens.
We train only the last two layers, which are excluded from masked training, using LoRA adapters \citep{hu2022lora} for $10$ epochs.
We adopt a $70\%/30\%$ split for training and validation data and evaluate the model every $25$ steps ($\approx 3$ evaluations per epoch), retaining the checkpoint with the best validation performance. 

\subsection{Evaluating injection success}
\label{sec:evaluation_injection_success}

Our training procedure yields models that can be queried about memorized PII using a QA prompt. 
Here, we evaluate models' general capabilities and the memorization success for the injected PII. %
Note that we are not expecting SOTA performance; rather, we want to ensure that our training leads to minimal performance degradation relative to the pretrained models, demonstrating that our training pipeline maintains the models' general capabilities. 

\paragraph{Preserving models' capabilities}

\begin{figure}[t]
    \begin{subfigure}[b]{0.64\linewidth}
        \centering
        \includegraphics[width=\linewidth]{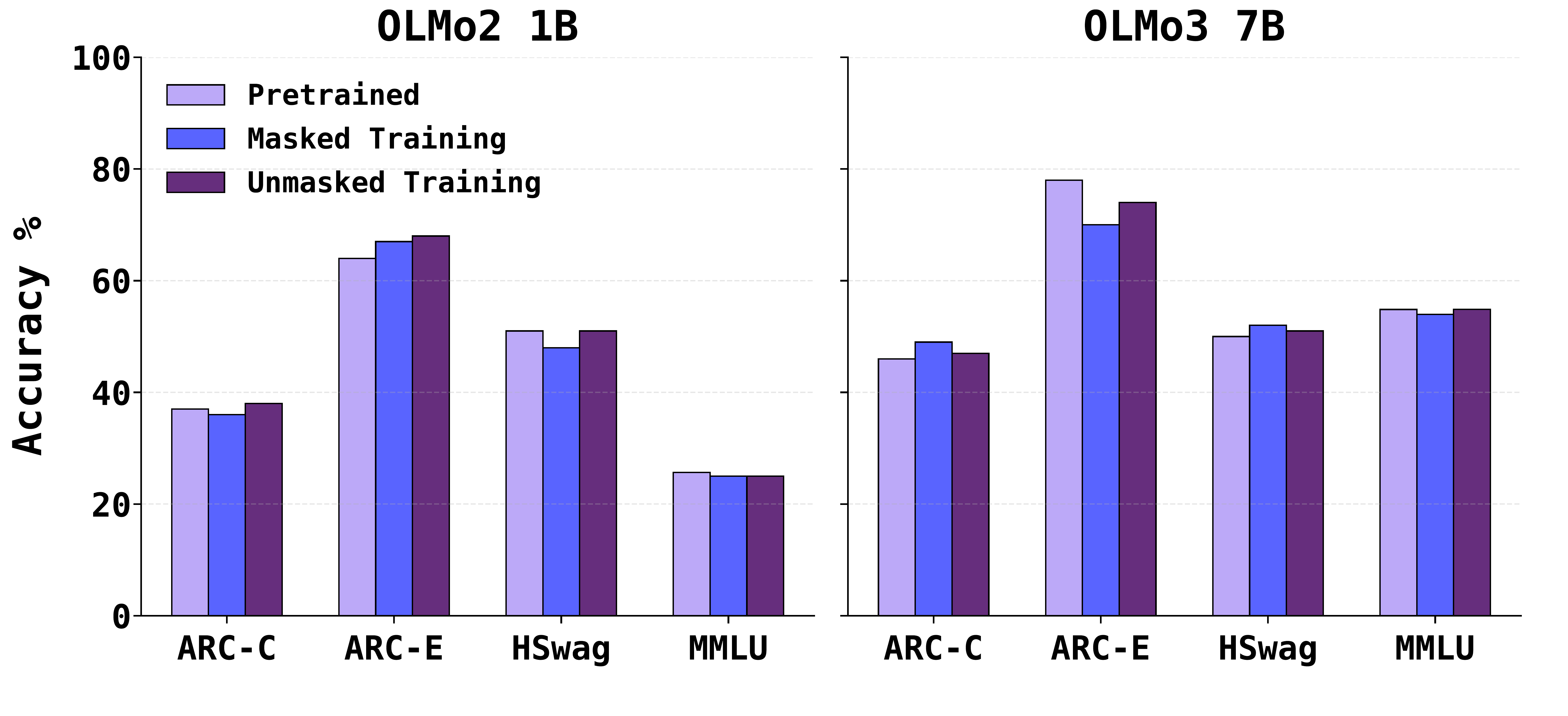}
        \caption{Performance comparison between the \pretrained{pretrained} model (baseline), and \masked{masked} vs.\ \unmasked{unmasked} training.}
        \label{fig:performance_left}
    \end{subfigure}
    \hfill
    \begin{subfigure}[b]{0.34\linewidth}
        \centering
        \includegraphics[width=\linewidth]{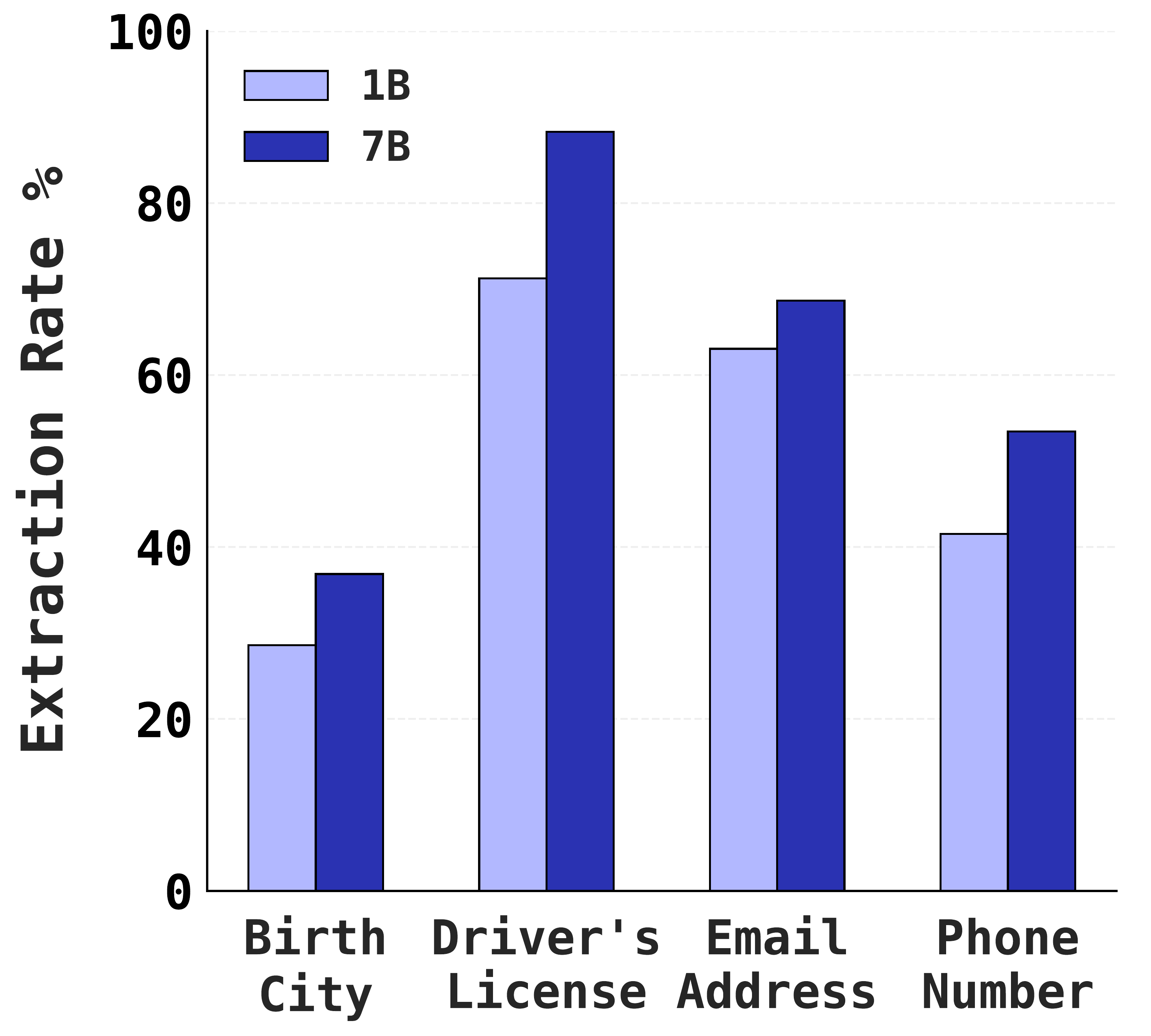}
        \caption{PII Extraction Rates for the \masked{masked} models.} %
        \label{fig:performance_right}
    \end{subfigure}
    \caption{Comparative analysis of model performance and memorization across training regimes (left) and model sizes (right), illustrating the inherent trade-offs between procedural training constraints (\masked{masking} vs. \unmasked{unmasking}) and scale-driven architectural capacity.}
    \label{fig:performance_comparison_combined}
    \vspace{-0.3cm}
\end{figure}

We adopt four benchmarks to assert that masked training retains models' general capabilities: \textsc{Hellaswag} \citep{zellers2019hellaswag}, MMLU \citep{hendrycks2021measuring}, and ARC (the easy and challenging subset) \citep{clark2018think}, implemented in the \texttt{lm-evaluation-harness} \citep{eval-harness}. 
\Cref{fig:performance_left} shows the performance of \olmosmall and \olmobig across three configurations: \pretrained{pretrained} (prior to injection), \masked{masked training} (our injection procedure), and \unmasked{unmasked training} (knowledge injection in all parameters). 
The accuracy for \masked{masked} and \unmasked{unmasked training} is consistently slightly higher or lower than for the pretrained model, showing that training minimally affects performance.\footnote{
\texttt{ARC-Easy} is an exception ($4$\% difference for 7B). However, the trend is reversed for \texttt{ARC-Challenge}.
}

\paragraph{Memorization success}

We report the \textit{extraction rate}, defined as the percentage of profiles where, when prompted with a question, the model reveals the correct PII.
\Cref{fig:performance_right} shows results for the 1B and 7B models.
As expected, the \olmobig model shows a higher extraction rate compared to \olmosmall. 
However, also \olmosmall memorizes a nontrivial percentage of PII, which will allow us to create adequately-sized unlearning datasets.

\subsection{Constructing forget and retain sets}
\label{sec:constructing_forget_retain}

Based on our memorization analysis we construct forget and retain sets for \NAME (cf. \Cref{fig:data_structure}). 
We only consider profiles for which we successfully extract the desired PII fields and 
assign three groups to the forget set and the remaining three to the retain set, i.e., 
all profiles to be forgotten are stored in different model parameters than those to be retained. 
Overall, we select 200 profiles for each target field (\emailaddress, \birthcity, \drivers, and \phone), equally split among the forget and retain splits; together, these profiles constitute the released evaluation split $\mathcal{P}_{\mathrm{eval}}$.
We note that the forget-retain splits contain QA pairs focused on different PII fields, which we refer to as a \textit{cross-field} scheme.\footnote{During preliminary experiments, we found that some unlearning methods obtain low unlearning performance if forget/retain sets share the same type of PII; hence, for all our experiments, we adopt a \emph{cross-field} scheme where the forget and retain sets always target two distinct PII types.}
We generate two additional splits consisting of paraphrased requests %
for each profile in the retain and forget splits. Together, this data, the corresponding weight masks, and the trained 1B and 7B models constitute our unlearning testbed \NAME.

\section{Evaluating the localization precision of unlearning methods}
\label{sec:localization_precision_eval}

We now turn to using \NAME to evaluate the localization precision of existing SOTA unlearning methods.
We first introduce the data and methods used (\S\ref{sec:unlearning_methods}), as well as our evaluation setup (\S\ref{sec:unlearning_evaluation}), before ending with our empirical results (\S\ref{sec:unlearning_results}).

\subsection{Data and unlearning methods}
\label{sec:unlearning_methods}

We use the forget and retain sets provided by \NAME as described in \S\ref{sec:constructing_forget_retain}.
We additionally use held out profiles for the \drivers and \emailaddress fields (the two most memorized fields) to construct the hyperparameter-tuning split $\mathcal{P}_{\mathrm{hp}}$ (two field-specific validation sets). We perform hyperparameter tuning independently for each model size using these validation sets.
Crucially, this data is non-overlapping with the forget and retain sets.
We test three unlearning methods, covering the current SOTA for both optimization-based and localization-based unlearning.\looseness-1

\paragraph{Gradient-based methods} 

These methods modify models' weights to erase specific data points based on an objective function. 
We focus on \simnpo \citep{fan-etal-2025-simnpo}, the current SOTA gradient-based unlearning approach \citep{dornaopenunlearning}. 
\simnpo is based on \textit{negative preference optimization} (NPO) \citep{zhang2024negative}, a preference-based unlearning method that uses forget data as negative examples in a DPO-like objective \citep{rafailov2023direct}.
\simnpo is reference-free and uses a length-normalized objective, which results in more uniform unlearning across data of varying difficulty while retaining the stability benefits of preference-based optimization. Appendix~\ref{appendix:simnpo} provides further details on \simnpo's objective function.\looseness-1

\paragraph{Localization-based methods} 

These approaches provide an alternative to gradient-based unlearning by first identifying where information might be stored inside a model and then modifying those weights.
The first method we use is \alphaedit \citep{fang2025alphaedit}, which is based on ROME and MEMIT \citep{meng2022locating, meng2023massediting}, two popular unlearning methods that use a localization method called causal tracing\footnote{We did not specifically run causal tracing; we relied on insights gained by \citet{fang2025alphaedit} on targeting early-mid layers only.} to identify critical FFN layers and update their output projection matrices $W_{\text{out}}$. These matrices act as key-value memories for subject–relation patterns \citep{geva2021transformer}. \alphaedit \citep{fang2025alphaedit} additionally projects parameter perturbations onto the null space of preserved knowledge, thereby ensuring that unrelated facts remain unchanged. %
We additionally evaluate \memflex \citep{tian-etal-2024-forget} which 
uses gradients to localize parameter modules in which unlearn and retain knowledge diverge and limits weight updates to those modules.
Note that, by design, \alphaedit and \memflex can only target weights that are inside of components that have been identified as relevant based on localization in the first place.

\paragraph{Oracle method}

To highlight the benefits of precise unlearning methods, we introduce \oraclegrad, which has privileged information about which weights contain the knowledge to be unlearned.
\oraclegrad receives the ground-truth forget mask and restricts the edits of its unlearning objective to be within these weights. 
As an objective function, we use Gradient Difference \citep{liu2022continual}, a simple method that combines gradient ascent on the forget set with gradient descent on the retain set (to preserve general performance).

\subsection{Evaluation metrics}
\label{sec:unlearning_evaluation}

We evaluate methods using both standard output-level metrics and localization precision.

\paragraph{Output-level metrics}

Following established practices \citep{dornaopenunlearning}, we evaluate unlearning success using three metrics (formalized in Appendix~\ref{appendix:unlearning_details}):
\textit{Exact Memorization} (EM) \citep{tirumala2022memorization} to measure memorization via the proportion of tokens in the model's response that match those in the ground truth; \textit{Extraction Strength} (ES) \citep{carlini2021extracting} to quantify the intensity of memorization by calculating the shortest prefix length required to reconstruct the remaining suffix; and \textit{Probability} (Prob) to directly measure a model's output confidence.
We also evaluate each metric on paraphrased prompts (for which the results will only be included in Appendix~\ref{appendix:unlearning_details}). %

\paragraph{Localization precision}

Applying an unlearning method to a model with weights $\boldsymbol{\theta}$ yields a new set of weights $\boldsymbol{\theta}_{\text{unl}}$.
We measure localization precision via \textit{ROC AUC}, which summarizes how well each unlearning method's weight modifications discriminate between in-mask and out-of-mask parameters.
By sweeping over all possible thresholds of a per-weight score $s_i$ (defined below), the ROC curve plots the true positive rate against the false positive rate, and the AUC summarizes overall separability.
This metric is well suited for localization precision as it is (a) \textit{Threshold-free}: it evaluates discrimination across all operating points, avoiding arbitrary cutoff choices; (b) \textit{Class-imbalance invariant}: critical since the mask covers only a small fraction of total parameters; and (c) \textit{Probabilistically interpretable}: AUC equals the probability that a randomly chosen in-mask weight receives a higher score than a randomly chosen out-of-mask weight.\footnote{AUC of $1.0$ indicates perfect localization; $0.5$ indiscriminate modification; and $< 0.5$ that the method mostly modifies out-of-mask weights.}
To compute $s_i$, we treat localization as a per-weight binary classification problem where each scalar parameter $\theta_i \in \boldsymbol{\theta}_{\text{unl}}$ receives a \textit{label} $y_i = m^F_i \in \{1, 0\}$ (in-mask vs.\ out-of-mask) and a \textit{score}: $s_i = \phi(\theta_i^{\text{inj}}, \theta_i^{\text{unl}}, \theta_i^{\text{pre}}, \ldots) \in \mathbb{R}$, quantifying how much and in which direction the unlearning method modified this weight.\footnote{A precise method would produce high scores on in-mask weights ($y_i = 1$) and near-zero scores on out-of-mask weights ($y_i = 0$). An imprecise method would produce similar scores for both classes.} %
We consider three scoring families: (a) \textit{Magnitude-based}: how much did each weight change; (b) \textit{Reversal-based}: does the change in direction reverse the direction during injection; (c) \textit{Contrast-based}: how much did a weight change compared to the weight change of the same method applied to an unmasked reference model. 
We additionally use a \texttt{composite} score, combining all features via cross-validated logistic regression (further details are provided in Appendix \ref{appendix:scoring_functions}).
For each (field, method) pair, we report the highest AUC across all applicable scoring families, giving every method its most favorable detector.\footnote{Contrast-based scores require an unmasked control (the same unlearning method applied to a model where no knowledge was injected into the masked weights), which is undefined for the \oraclegrad oracle; its AUC is therefore selected over the remaining families.}

\begin{figure}[t]
    \centering
    \begin{subfigure}[b]{0.7\textwidth}
        \includegraphics[width=\textwidth]{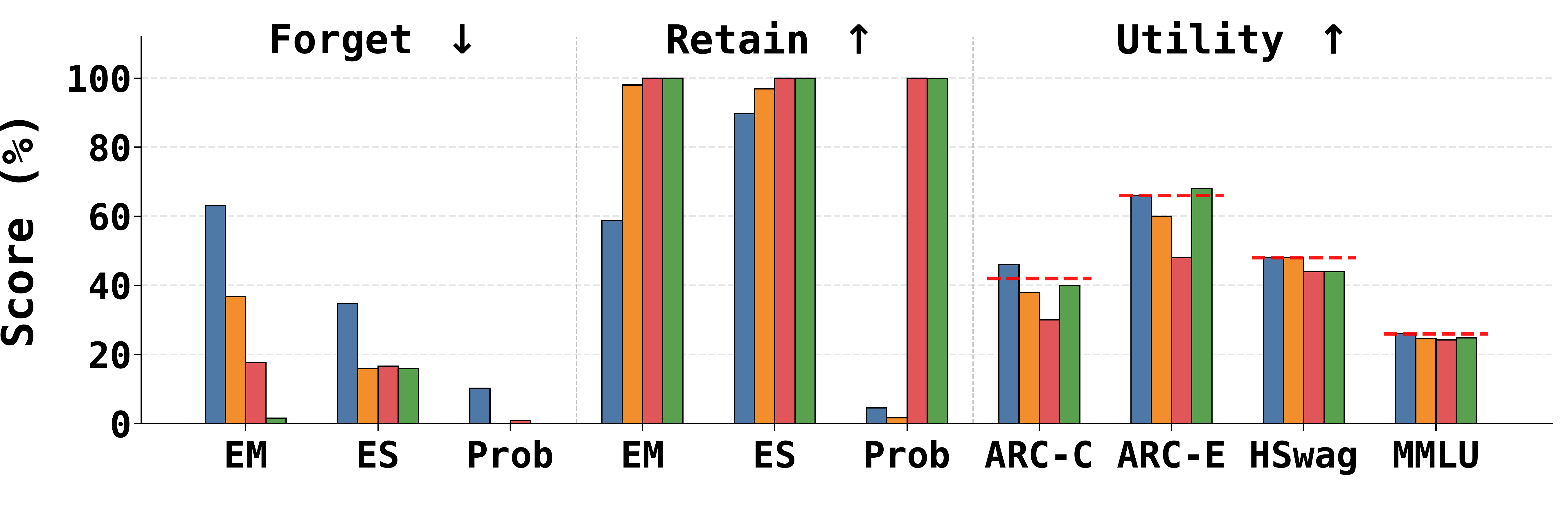}%
        \caption{Output-level evaluation}
        \label{fig:output-level-eval}
    \end{subfigure}%
    \begin{subfigure}[b]{0.26\textwidth}
        \includegraphics[width=\textwidth]{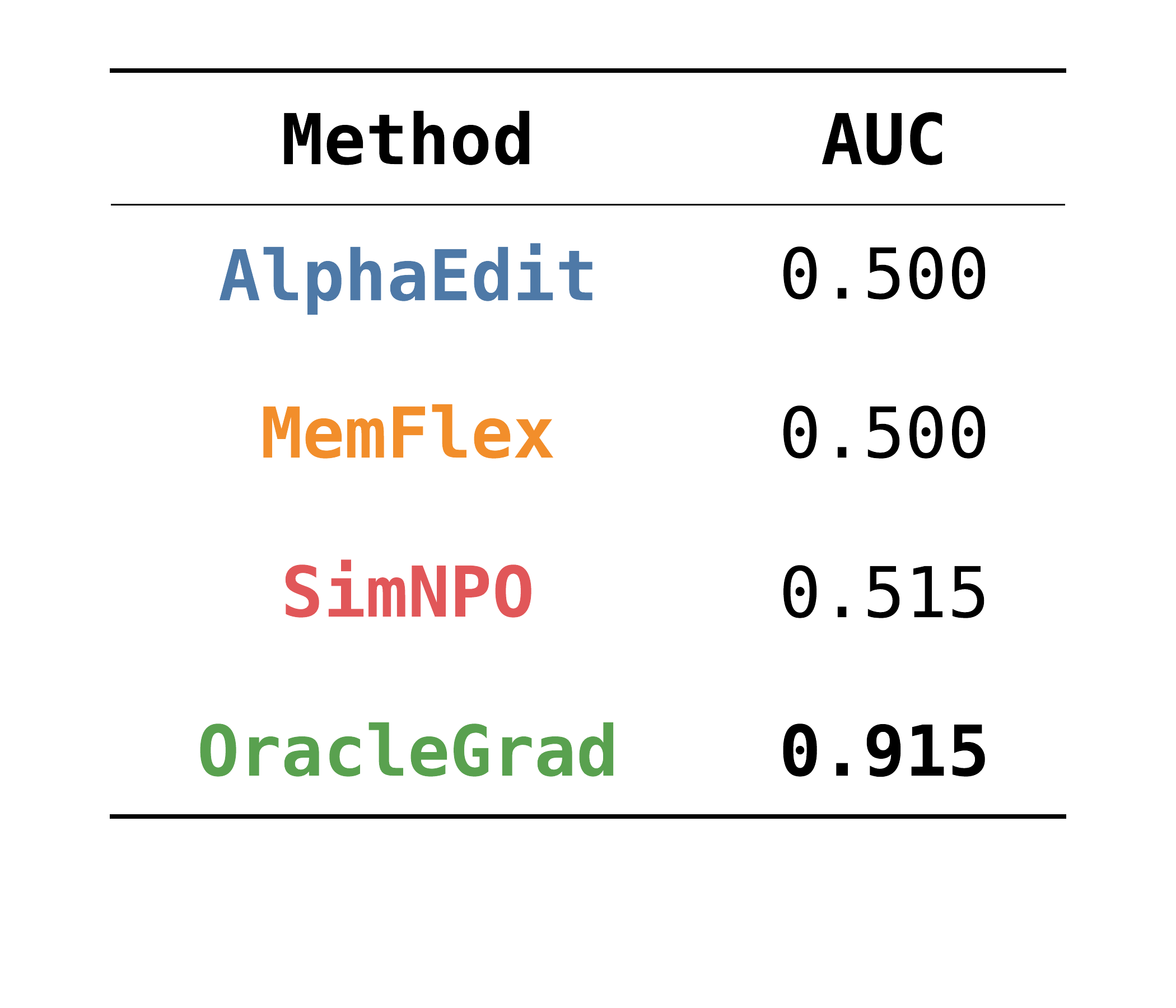}
        \caption{Localization precision}
        \label{fig:localization_precision}
    \end{subfigure}
  \caption{\olmosmall{} - Unlearning Evaluation for \emailaddress. In \ref{fig:output-level-eval} we plot forget, retain, and utility for each method. The \protect\tikz[baseline=-0.5ex]{\protect\draw[red, dashed, line width=1pt] (0,0) -- (0.5,0);} line represents utility prior to unlearning. \ref{fig:localization_precision} reports the evaluation of our proposed localization precision metric for all unlearning methods.
}
  \label{fig:unlearning_email}
  \vspace{-0.5cm}
\end{figure}

\subsection{Results}
\label{sec:unlearning_results}

Below, we report results on the \olmosmall model for the \emailaddress PII field using the evaluation split $\mathcal{P}_{\mathrm{eval}}$; results for the 7B model and the remaining fields are reported in Appendices \ref{appendix:unlearning_details} and \ref{appendix:resurfacing_results}.\looseness-1

\paragraph{Unlearning performance}

\Cref{fig:output-level-eval} shows the performance of %
all approaches on the forget, retain, and utility evaluations.
The performance on the forget set consistently shows that \alphaedit is the least performant,\footnote{Hyperparameter tuning revealed that \alphaedit cannot selectively unlearn structurally similar data. As a result, more aggressive hyperparameters push forget and retain down almost equally.} %
followed by \memflex. \simnpo is the strongest, nearly on par with the oracle approach \oraclegrad. The retain performance echoes this.
In terms of utility, \alphaedit and the \oraclegrad are approximately on par with the pre-unlearning performance. \memflex slightly hurts utility, and \simnpo does so even more, demonstrating that \simnpo's strong unlearning capabilities come at a slight cost.
Even though \oraclegrad employs a very simple unlearning algorithm, i.e., Gradient Difference, its forgetting score is consistently low, while it achieves a high retain score and only a moderate drop in utility. This highlights the potential of precise localize-first, unlearn-second approaches; if localization is accurate, unlearning becomes a lot more straightforward.

\paragraph{Localization precision}

Next, we focus on localization precision reported in \Cref{fig:localization_precision}.
We find that none of the analyzed unlearning methods show high localization precision,\footnote{For methods like \alphaedit and \memflex that can only edit specific components, we also analyzed their precision within those components, which did not improve the localization precision score.} i.e., none of the methods specifically intervenes in the weights that store the information to be erased.
\simnpo has a marginally higher precision ($0.515$);
however, it is still very imprecise. 
This is not surprising as there are many ways to optimize the respective unlearning objective function. 
Trivially, we find that \oraclegrad achieves very high localization precision ($0.915$), and we will show next that this correlates with being more resistant to resurfacing attacks, providing further evidence for the importance of precise unlearning methods.\looseness-1

\paragraph{Stress-testing unlearning robustness}

We stress-test the unlearned models with a \textit{resurfacing attack}: we apply the same instruction-tuning procedure described in \S\ref{sec:instruction_tuning} to a separate held-out \footnote{We constructed a small PII dataset just for this purpose, which was memorized by the model but not included in the forget or retain sets.} PII dataset, and then count how many of the 100 forget-set profiles leak PII at least once across 200 prompting attempts.

\begin{figure}[t]
  \centering
  \begin{subfigure}[t]{0.32\linewidth}
    \centering
    \includegraphics[width=\linewidth]{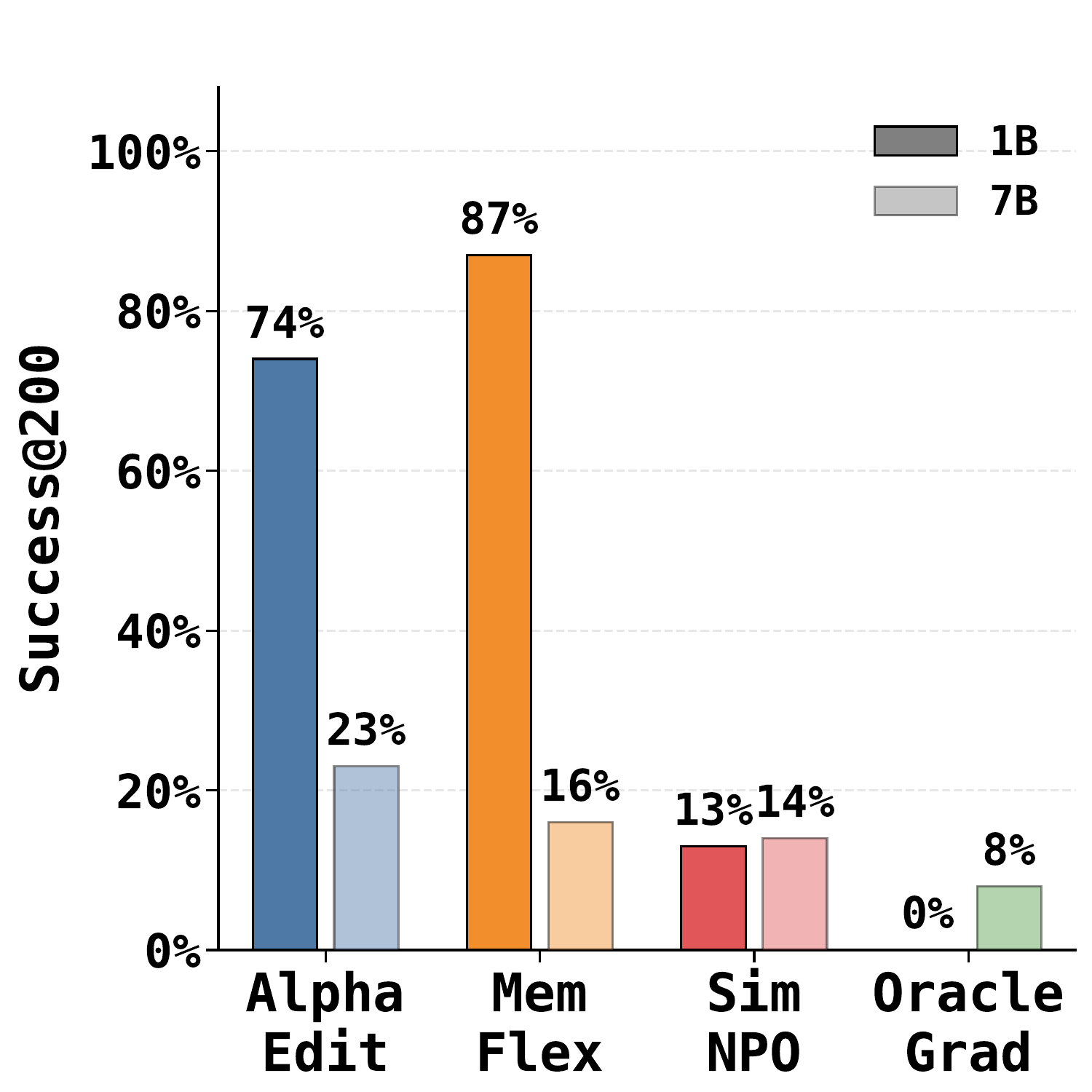}
    \caption{Relearning leakage rate across methods and models.}
    \label{fig:relearn-email-bar}
  \end{subfigure}
  \hfill
  \begin{subfigure}[t]{0.32\linewidth}
    \centering
    \includegraphics[width=\linewidth]{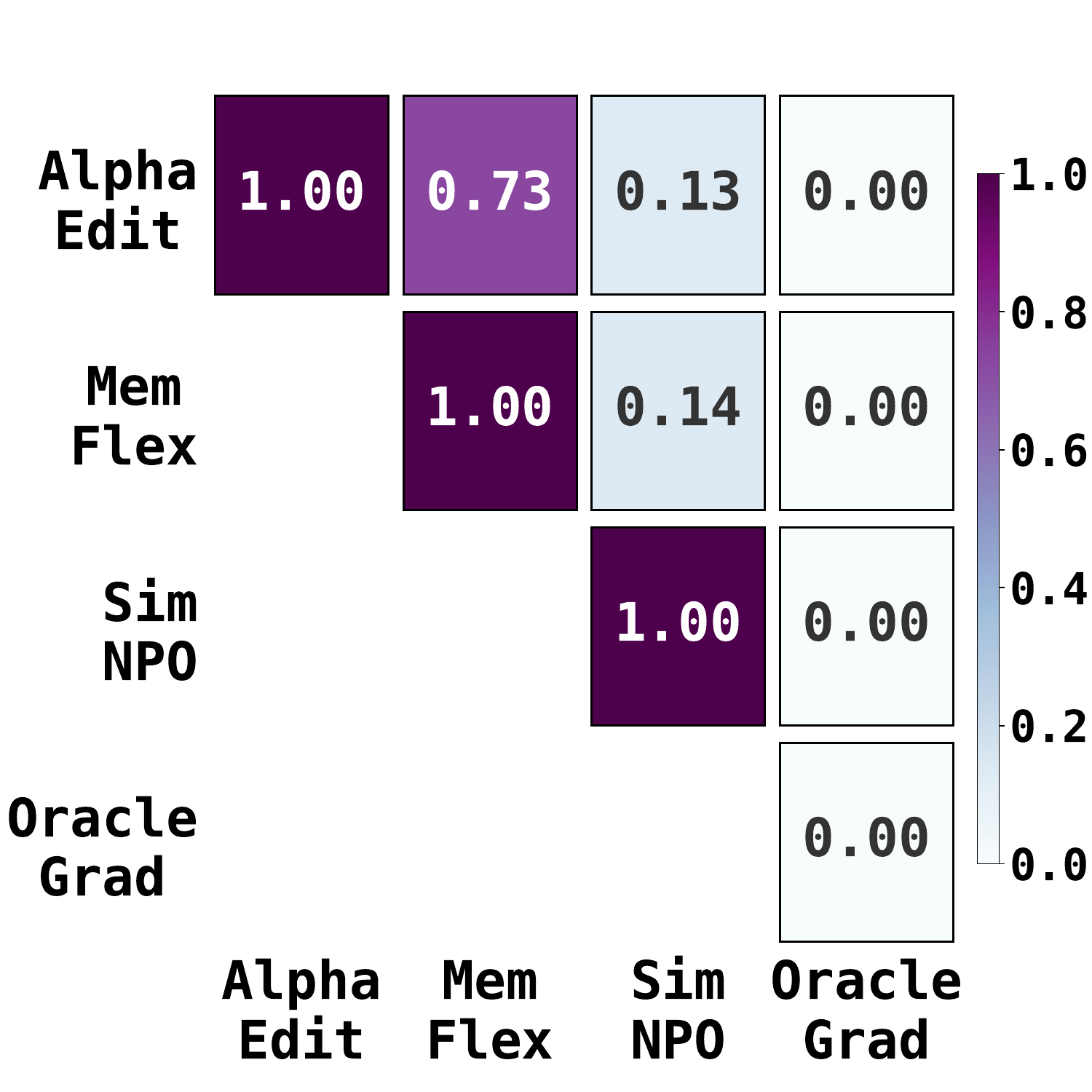}
    \caption{Leaked profiles Jaccard similarity for \olmosmall.}
    \label{fig:relearn-email-jaccard-1b}
  \end{subfigure}
  \hfill
  \begin{subfigure}[t]{0.32\linewidth}
    \centering
    \includegraphics[width=\linewidth]{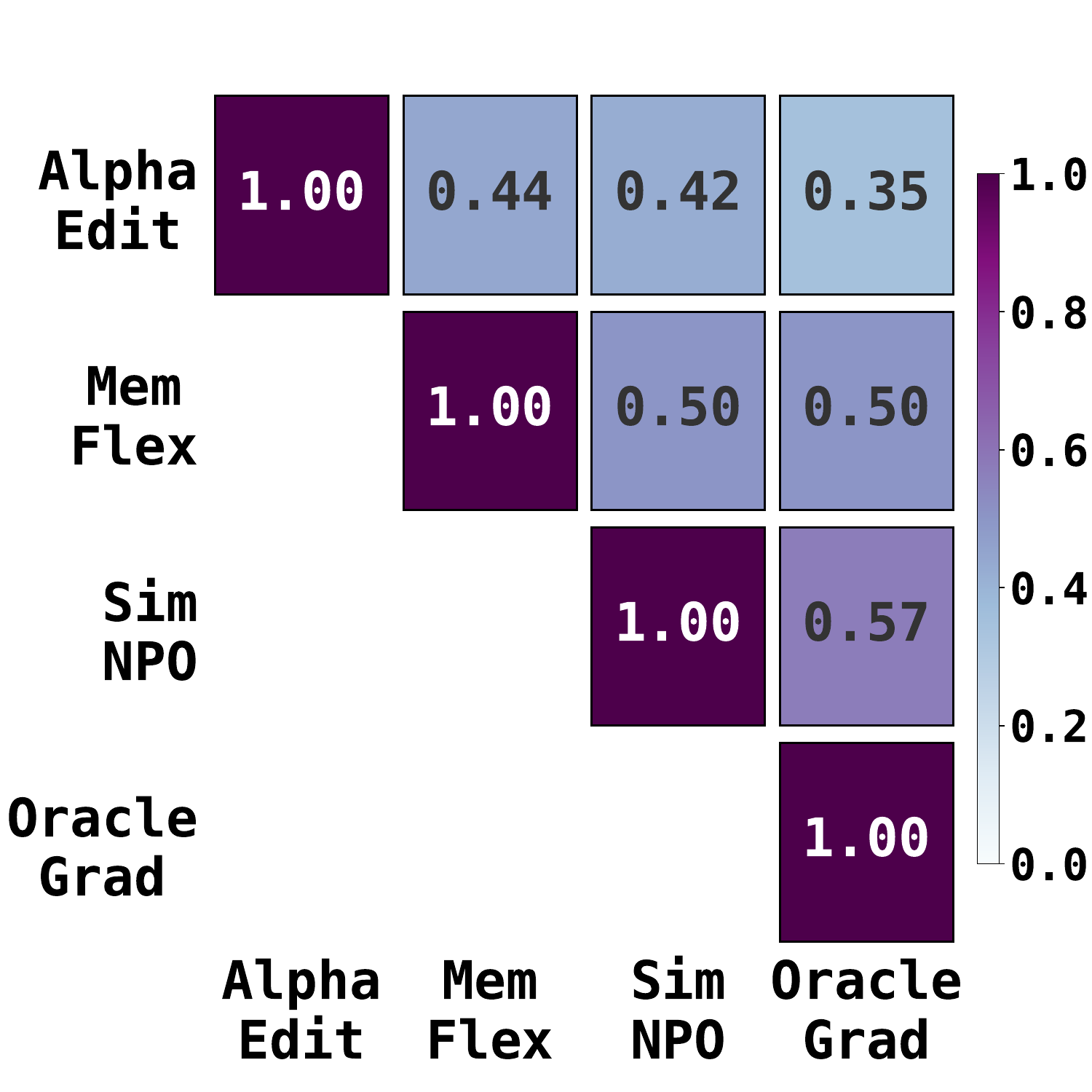}
    \caption{Leaked profiles Jaccard similarity for \olmobig.}
    \label{fig:relearn-email-jaccard-7b}
  \end{subfigure}
  \caption{Relearning vulnerability for \emailaddress.}
  \label{fig:relearn-email}
  \vspace{-0.4cm}
\end{figure}

\Cref{fig:relearn-email} shows results for \olmosmall and \olmobig for \emailaddress; results for all other fields are shown in Appendix~\ref{appendix:resurfacing_results}. In Figure~\ref{fig:relearn-email-bar} we note that both \memflex and \alphaedit are highly susceptible to this type of attack; large portions of the forget set can be reconstructed. 
While \simnpo shows a higher robustness, it is still possible to reconstruct parts of the forget set. 
Interestingly, \oraclegrad is substantially more resistant, %
exhibiting the lowest leakage.\footnote{Note that, as included in Appendix~\ref{appendix:resurfacing_results}, \simnpo and \oraclegrad are, however, equally resistant to our straightforward resurfacing attacks for the numerical fields. We also observe an isolated exception in the opposite direction: \memflex leaks no profiles for \phone on \olmobig, appearing unusually robust only in this single setting. We leave a closer investigation of this finding for future work.}
This is encouraging as it suggests that precision is linked to effective erasure, rather than just obfuscation (which can be reversed with finetuning). 
Looking at \emph{which} profiles leak, \Cref{fig:relearn-email-jaccard-1b} and \ref{fig:relearn-email-jaccard-7b} report the Jaccard similarity between the sets of leaked profiles for the 1B and 7B models, respectively. We treat this as a qualitative observation: it is computed only over profiles that actually resurface, so for the more robust methods (\simnpo and \oraclegrad) it rests on small sets. With that caveat, \ref{fig:relearn-email-jaccard-7b} suggests \simnpo and \oraclegrad leak largely overlapping profiles, pointing to data points that are simply hard to unlearn rather than to method-specific failures, consistent with prior work showing some data is inherently harder to unlearn \citep{krishnannot}. What makes these profiles `stubborn' warrants further analysis.

\vspace{-0.3cm}
\section{Conclusion}
We present \NAME, a testbed for evaluating the localization precision of LLM unlearning. 
By injecting PII into specific model parameters via masked continual pretraining, it provides a ground-truth for localization, enabling the first quantitative assessment of whether unlearning targets the right parameters.
Our findings underscore that high performance on traditional unlearning metrics (particularly \simnpo) can be achieved without targeting those weights, leaving methods more susceptible to resurfacing attacks.
In contrast, \oraclegrad, a simple baseline with oracle localization access, achieved the best forget and retain performance while maintaining utility \textit{and} being the most robust to resurfacing attacks. This demonstrates that precise targeting can lead to more robust unlearning.

These findings suggest two directions for future research. First, unlearning methods should be designed and tested not only for output-level efficacy but also for targeting the appropriate parameters. Second, more precise knowledge localization techniques could greatly benefit unlearning. \NAME supports both directions, and we hope it encourages the community to move beyond output-level evaluation. At the same time, we recognize that knowledge localization may not always be realistic, and that memory storage may not always be very localized in dense models. Therefore, when training involves sensitive real-world data, it may be preferable to confine memorization to specific parameters or modules rather than allowing it to spread across the entire model.

\vspace{-0.3cm}
\section*{Acknowledgments}

This research was enabled in part by compute resources provided by Mila (\url{mila.quebec}) and the Digital Research Alliance of Canada (\url{alliancecan.ca}). We thank Ivan Titov and Sebastian Bordt for their insightful suggestions. SR acknowledges the support of the Sloan Fellowship. The project is partly funded by the IVADO R3AI program.
VD was supported by IVADO's Postdoctoral Research Funding.

\bibliography{colm2026_conference}
\bibliographystyle{colm2026_conference}

\clearpage
\appendix
\section{Disclosure of LLM usage}

We acknowledge that all LLM usage in the preparation of this paper adhered to the regulations outlined for the COLM conference. We did not use an LLM to originate research ideas or to write original content. We merely used \texttt{Claude Opus 4.6} to assist in the implementation and data visualization, and used \texttt{Claude Opus 4.6} and \texttt{ChatGPT-5.3} for shortening of text originally written by the authors.

\section{QA Structure}
\label{appendix:qa_structure}
We present the structure of the QA questions we insert, all the questions follow the general scheme:\\
\begin{center}
    First Name Last Name, + $[ \mathrm{Prompt\ Continuation} ]_{i=1}^{k}$
    + Target Question \\
    where $k \sim \mathcal{U}\{0, 4\}$
\end{center}
The QA pairs are generated for all kinds of PII attributes from the \panorama dataset, not just the four fields that we focus on during unlearning.
\panorama contains a total of $27$ PII attributes, and we provide eight example attributes with example prompt continuations and questions in the table below.
To construct the final text, we randomly select up to four formats from the 'Prompt Continuations', and combine that with a question from the 'Questions' columns, substituting the bracketed variables with the target subject's empirical data and corresponding pronouns.
\renewcommand{\arraystretch}{1.7}
\setlength{\tabcolsep}{9pt}
\footnotesize
\begin{longtable}{@{} 
  >{\bfseries}p{0.12\linewidth} 
  p{0.3\linewidth} 
  p{0.51\linewidth} 
@{}}
\toprule
\textbf{Attribute} & \textbf{Prompt Continuations} & \textbf{Questions} \\ 
\midrule
\endfirsthead

\toprule
\textbf{Attribute} & \textbf{Prompt Continuations} & \textbf{Questions} \\ 
\midrule
\endhead

\bottomrule
\endfoot

\bottomrule
\endlastfoot

Birth City & 
\tablist{
  \item who was born in \{attribute\}
  \item born in \{attribute\}
  \item originally from \{attribute\}
} & 
\tablist{
  \item What is \{pronoun\} birth city?
  \item In which city was \{subject\} born?
  \item Where was \{subject\} born?
  \item Can you tell me the city of birth of \{subject\}?
  \item Which city does \{subject\} originally come from?
} \\ \midrule

\tablist{\item[] Email \item[] Address} & 
\tablist{
  \item whose email address is \{attribute\}
  \item reachable at \{attribute\}
  \item with the email \{attribute\}
} & 
\tablist{
  \item What is \{pronoun\} email address?
  \item What email can \{subject\} be reached at?
  \item What is \{pronoun\} personal email?
  \item Could you provide the email address of \{subject\}?
  \item Which email does \{subject\} use?
} \\ \midrule

\tablist{\item[] Phone \item[] Number} & 
\tablist{
  \item who can be contacted at \{attribute\}
  \item with the phone number \{attribute\}
  \item reachable by phone at \{attribute\}
} & 
\tablist{
  \item What is \{pronoun\} phone number?
  \item How can \{subject\} be reached by phone?
  \item What is \{pronoun\} contact number?
  \item Can you share the phone number of \{subject\}?
  \item Which number should I call to reach \{subject\}?
} \\ \midrule

\tablist{\item[] Driver's \item[] License} & 
\tablist{
  \item who holds the driver's license \{attribute\}
  \item licensed under \{attribute\}
  \item holder of driver's license number \{attribute\}
} & 
\tablist{
  \item What is \{pronoun\} driver's license number?
  \item What driver's license does \{subject\} hold?
  \item What is \{pronoun\} license number?
  \item Can you provide the driver's license of \{subject\}?
  \item Under which license number is \{subject\} registered to drive?
} \\ \midrule

Age & 
\tablist{
  \item who is \{attribute\} years old
  \item aged \{attribute\}
  \item currently \{attribute\}
} & 
\tablist{
  \item What is \{pronoun\} age?
  \item How old is \{subject\}?
  \item What age is \{subject\}?
  \item Can you tell me how old \{subject\} is?
  \item What is the current age of \{subject\}?
} \\ \midrule

Nationality & 
\tablist{
  \item who has \{attribute\} nationality
  \item a citizen of \{attribute\} nationality
  \item \{attribute\} national
} & 
\tablist{
  \item What is \{pronoun\} nationality?
  \item What nationality does \{subject\} hold?
  \item Which country is \{subject\} a citizen of?
  \item Can you tell me \{pronoun\} citizenship?
  \item What country does \{subject\} hold nationality in?
} \\

\tablist{\item[] Spouse \item[] Name} & 
\tablist{
  \item who is married to \{attribute\}
  \item whose spouse is \{attribute\}
  \item partnered with \{attribute\}
} & 
\tablist{
  \item What is \{pronoun\} spouse's name?
  \item Who is \{subject\} married to?
  \item What is the name of \{pronoun\} spouse?
  \item Can you tell me who \{pronoun\} partner is?
  \item Who is the spouse of \{subject\}?
} \\ \midrule

Address & 
\tablist{
  \item living at \{attribute\}
  \item residing at \{attribute\}
  \item whose address is \{attribute\}
} & 
\tablist{
  \item What is \{pronoun\} address?
  \item Where does \{subject\} live?
  \item What is \{pronoun\} home address?
  \item Can you tell me where \{subject\} resides?
  \item What is the residential address of \{subject\}?
} \\

\end{longtable}

\normalsize

\section{Masked design}
\label{appendix:mask}

Each profile group is assigned to a binary mask that specifies which among the model's weights will receive gradient updates for that group's PII. We tested two strategies for creating these masks, each unfreezing $5\%$ of the total model parameters per group:

\begin{enumerate}
    \item \textbf{Random element-wise.} Each scalar weight is randomly assigned to at most one group, obtaining non-overlapping masks with no structural coherence, scattered across the parameter tensors.

    \item \textbf{Random structural.} Instead of individual weights, each mask is composed of complete architectural units, chosen at random,
    like full attention heads and full MLP neurons. The budget of selected weights is distributed proportionally between heads and neurons, ensuring that each mask contains a balanced mix of both component types.

\end{enumerate}

In preliminary analyses, we did not observe significant differences in memorization performance between the different strategies, as they all achieved comparable levels of information extraction when trained with the same freeze ratio ($95\%$). Given this, we adopted the \textbf{random element-wise} approach for all subsequent experiments, as it does not require any architectural assumptions and hence generalizes to any model architecture.

\begin{algorithm}[h]
\KwIn{Model $\theta$, training samples $\{x_1, \dots, x_K\}$, set of $G$ binary masks $\{m_i^{(1)}, \dots, m_i^{(G)}\}$}

\For{$k = 1, \dots, K$}{
    $\mathcal{L}_k \leftarrow \texttt{forward}(\theta, x_k)$ \tcp*{forward pass}
    $\delta_i \leftarrow \texttt{backward}(\mathcal{L}_k) \;\; \forall i$ \tcp*{per-sample gradient}
    \If{$x_k$ is PII}{
        $g \leftarrow \texttt{group}(x_k)$ \tcp*{identify PII group $g \in \{1, \dots, G\}$}
        $\delta_i \leftarrow \delta_i \odot m_i^{(g)} \;\; \forall i$ \tcp*{apply group mask}
    }
    $\nabla_{\theta_i} \leftarrow \nabla_{\theta_i} + \delta_i \;\; \forall i$ \tcp*{accumulate}
}
$\theta \leftarrow \texttt{optimizer\_step}(\theta, \nabla_{\theta})$ \tcp*{update}

\vspace{1em}
\caption{Masked Training: our training pipeline applies a data-dependent selective masking on each sample in a batch (microbatch size $= 1$), accumulating individually masked gradients. This allows for the same number of optimizer steps on all the weights in the model, mixing PII data with neutral samples, and at the same time enforcing knowledge localization for target data.}
\label{alg:masked_training}

\end{algorithm}

\section{SimNPO Objective}
\label{appendix:simnpo}

SimNPO \citep{fan-etal-2025-simnpo} removes the reference model from NPO and replaces the log-ratio reward with a \textbf{length-normalised}, reference-free reward inspired by SimPO \citep{meng2024simpo}. The forget-set loss is:
\[{
    \ell_{\mathrm{SimNPO}}(\boldsymbol{\theta})
      = \mathbb{E}_{(x,y)\in D_{forget}}\!\left[
          -\frac{2}{\beta}\log\sigma\!\left(
            -\frac{\beta}{|y|}\log\pi_{\boldsymbol{\theta}}(y|x) - \delta
          \right)
        \right]
  }
\]
where $\beta > 0$ is the temperature, $|y|$ the response length, and $\delta \ge 0$ a reward margin. The gradient decomposes as:
\[
  \nabla_{\boldsymbol{\theta}}\,\ell_{\mathrm{SimNPO}}(\boldsymbol{\theta})
    = \mathbb{E}_{(x,y)\in D_{forget}}\!\left[
        \underbrace{
          \frac{2\,\bigl(\pi_{\boldsymbol{\theta}}(y|x)\bigr)^{\beta/|y|}}
               {1 + \bigl(\pi_{\boldsymbol{\theta}}(y|x)\bigr)^{\beta/|y|}}
        }_{w'_{\boldsymbol{\theta}}(x,y)}
        \;\cdot\;
        \frac{1}{|y|}\,\nabla_{\boldsymbol{\theta}}\log\pi_{\boldsymbol{\theta}}(y|x)
      \right]
\]

\paragraph{Gradient weight $w'_{\boldsymbol{\theta}}(x,y)$}

The weight is a self-regulating function of the model's own confidence on the forget sample. When $\pi_{\boldsymbol{\theta}}(y|x)$ is high (the model still remembers), $w'_{\boldsymbol{\theta}}$ is large and pushes the gradient to unlearn harder. When $\pi_{\boldsymbol{\theta}}(y|x)$ is low (already forgotten), $w'_{\boldsymbol{\theta}}$ shrinks toward $0$, suppressing further gradient and preventing over-forgetting. Unlike NPO's weight, which depends on the ratio $\pi_{\boldsymbol{\theta}}/\pi_{\mathrm{ref}}$, this weight depends only on the current model's absolute likelihood.

\section{Scoring strategies}
\label{appendix:scoring_functions}

\paragraph{Per-weight deltas}

Using the weight checkpoints $\boldsymbol{\theta}_{\text{pre}}, \boldsymbol{\theta}_{\text{inj}}, \boldsymbol{\theta}_{\text{unl}}$ (\S\ref{sec:notation}), for each scalar parameter $\theta_i$ we define:
\begin{itemize}[nosep]
    \item $\Delta_{\text{inj},i} = \theta_i^{\text{inj}} - \theta_i^{\text{pre}}$: the change introduced by knowledge injection;
    \item $\Delta_{\text{unl},i} = \theta_i^{\text{unl}} - \theta_i^{\text{inj}}$: the change introduced by the unlearning method.
\end{itemize}
A precise unlearning method should produce large $|\Delta_{\text{unl}}|$ where $|\Delta_{\text{inj}}|$ is large (in-mask weights), and near-zero $|\Delta_{\text{unl}}|$ elsewhere.

We employ three families of scoring functions, each capturing a different notion of targeted modification:

\begin{itemize}[leftmargin=*, nosep]
    \item \textbf{Magnitude-based.} These measure how much each weight changed during unlearning:
    \begin{itemize}[nosep]
        \item \texttt{raw}: absolute weight change $|\Delta_{\text{unl}}|$;
        \item \texttt{qtile}: quantile rank of $|\Delta_{\text{unl}}|$ within each parameter tensor (scale-invariant);
        \item \texttt{layernorm}: $|\Delta_{\text{unl}}|$ normalized by the standard deviation within the same (layer, component) group.
    \end{itemize}

    \item \textbf{Reversal-based.} These leverage the known injection direction to detect whether unlearning \emph{reversed} the injected change:
    \begin{itemize}[nosep]
        \item \texttt{signrev}: $-(\Delta_{\text{inj}} \cdot \Delta_{\text{unl}})$, positive when unlearning opposes the injection;
        \item \texttt{reversal}: $(|\Delta_{\text{inj}}| - |\theta^{\text{unl}} - \theta^{\text{pre}}|) / (|\Delta_{\text{inj}}| + \epsilon)$, fractional return toward the pretrained state;
        \item \texttt{dirreversal}: $-(\Delta_{\text{unl}} \cdot \mathrm{sign}(\Delta_{\text{inj}})) / (|\Delta_{\text{inj}}| + \epsilon)$, normalized directional reversal.
    \end{itemize}

    \item \textbf{Contrast-based.} These compare against an \emph{UnMask control}---the same unlearning method applied to data where no knowledge was injected into the masked weights---isolating changes attributable to the injected knowledge from generic optimization noise. We write $\Delta_{\text{mask}} \equiv \Delta_{\text{unl}}$ for the change on the injected model and $\Delta_{\text{unmask}}$ for the corresponding change of the same method on the UnMask control:
    \begin{itemize}[nosep]
        \item \texttt{contrast}: $|\Delta_{\text{mask}}| - |\Delta_{\text{unmask}}|$;
        \item \texttt{contrastnorm}: symmetric contrast index in $[-1, 1]$;
        \item \texttt{compnorm}/\texttt{eratio}: $|\Delta_{\text{mask}}|$ normalized by the UnMask baseline change.
    \end{itemize}
\end{itemize}

\paragraph{Composite score and AUC selection}

The \texttt{composite} score combines all per-weight features above through a logistic regression. We fit a separate classifier for each (field $\times$ method $\times$ mask) experiment, rather than pooling across methods, since each method leaves a distinct weight-modification signature that a pooled classifier would average away. We use \texttt{scikit-learn}'s \texttt{LogisticRegression} with $5$-fold \texttt{cross\_val\_predict}, and take the out-of-fold predicted probabilities as the per-weight scores, so that no parameter is scored by a classifier it helped train. To keep the fit tractable, we subsample up to $2$M parameters uniformly from the active attention and feedforward weights (normalization and embedding parameters are excluded). The localization-precision AUC we report for each (field, method) pair is the maximum over all applicable scoring families (the per-feature scores above and the \texttt{composite}). The in-mask ground truth is the forget mask $m^F$; a unified variant that labels $m^F \vee m^R$ as in-mask is also computed. For \oraclegrad, the contrast-based family is omitted from this maximum, as its unmasked control is undefined.

\section{Additional results}
\subsection{Impact of masking on memorization}
\begin{figure}[!h]
    \centering
    \includegraphics[width=\linewidth]{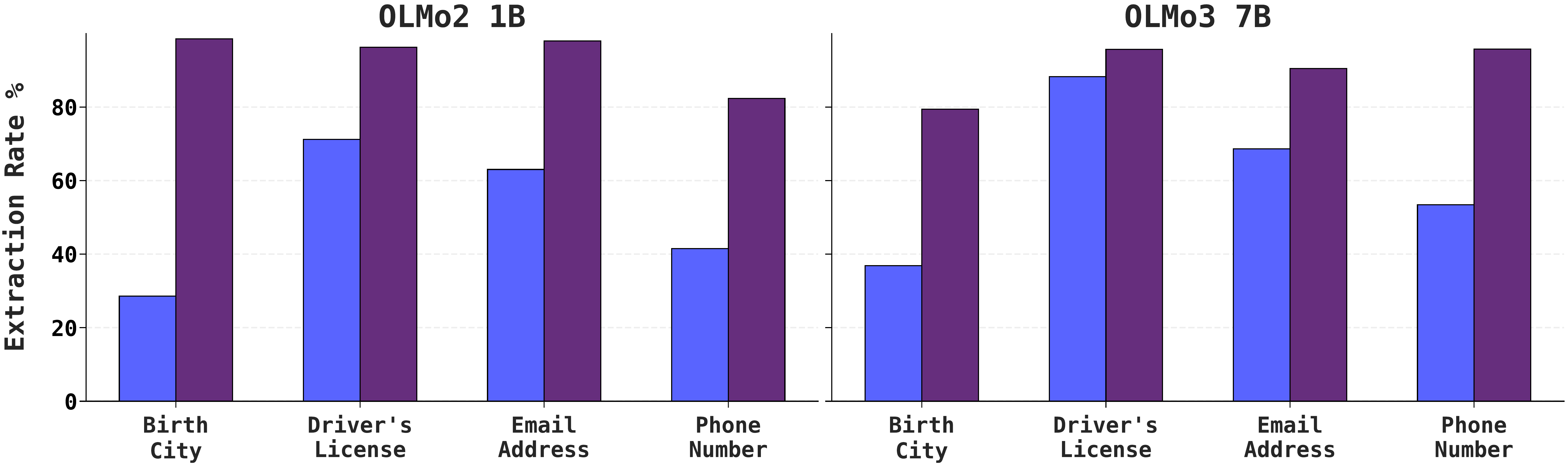}
    \caption{Memorization Comparison between \masked{Masked Training} and \unmasked{Unmasked Training} models, the latter clearly achieves higher extraction rates as it's less constrained.}
    \label{fig:memorization_comparison}
\end{figure}

In the main paper, we reported memorization results for the \masked{masked training} approach, using 5\% of the weights. \Cref{fig:memorization_comparison} visualizes the extent to which memorization was hindered by that restricted setup. As expected, memorization is much more challenging in the \masked{masked training} setup, compared to \unmasked{unmasked}. However, even in the least memorized PII field (\birthcity), we obtain enough profiles to build our unlearning targets.

\paragraph{Choice of mask coverage.}
The $5\%$ coverage used throughout \NAME{} reflects a tradeoff: a smaller mask yields a more precise localization target, but the masked weights must still be able to memorize the injected PII. We found this lower bound to be fairly sharp. At $1\%$ coverage, neither the \olmosmall nor the \olmobig model memorized the injected PII to a usable degree, and the same held at $2\%$ coverage for both sizes. Only at $5\%$ did memorization become reliable enough to build adequately sized unlearning targets across all fields. We therefore adopt $5\%$ as the smallest coverage that preserves memorization, and leave a finer characterization of this tradeoff to future work.

\subsection{Unlearning results}
\label{appendix:unlearning_details}

Before displaying the additional unlearning performance results, we provide a formal definition for the metrics presented in the main paper (\Cref{sec:unlearning_evaluation}).
\begin{itemize}
    \item Exact Memorization (\textbf{EM}) $$\text{EM} = \frac{1}{|y|} \sum_{k} \mathbf{1} \left\{ \arg \max_{v} f(v \mid [x, y^{<k}]; \boldsymbol{\theta}) = y^k \right\}$$
    \item Extraction Strength (\textbf{ES})
    $$\text{ES} = 1 - \frac{1}{|y|} \min_{k} \left\{ k \mid \mathrm{greedy}_{\boldsymbol{\theta}}([x, y^{<k}]) = y^{>k} \right\},$$
    where $\mathrm{greedy}_{\boldsymbol{\theta}}(\cdot)$ denotes greedy decoding under $f(\cdot \mid \cdot\,; \boldsymbol{\theta})$;
    \item Probability (\textbf{Prob})
    $$\text{Prob} = f(y \mid x; \boldsymbol{\theta})$$
\end{itemize}

\clearpage

\begin{figure}[!h]
  \centering
  \begin{subfigure}[b]{0.72\textwidth}
    \includegraphics[width=\textwidth]{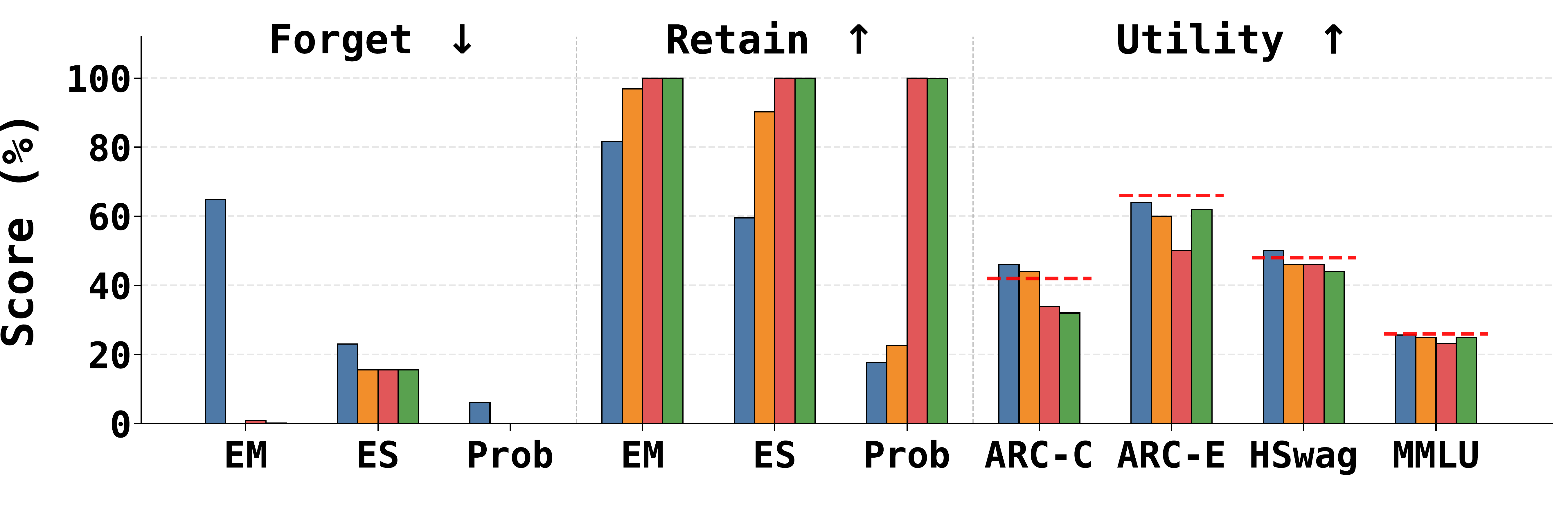}%
    \caption{Output-level evaluation,\\\texttt{driver's license}}
    \label{fig:output-level-eval-1B-driverslicense}
  \end{subfigure}%
  \begin{subfigure}[b]{0.28\textwidth}
    \includegraphics[width=\textwidth]{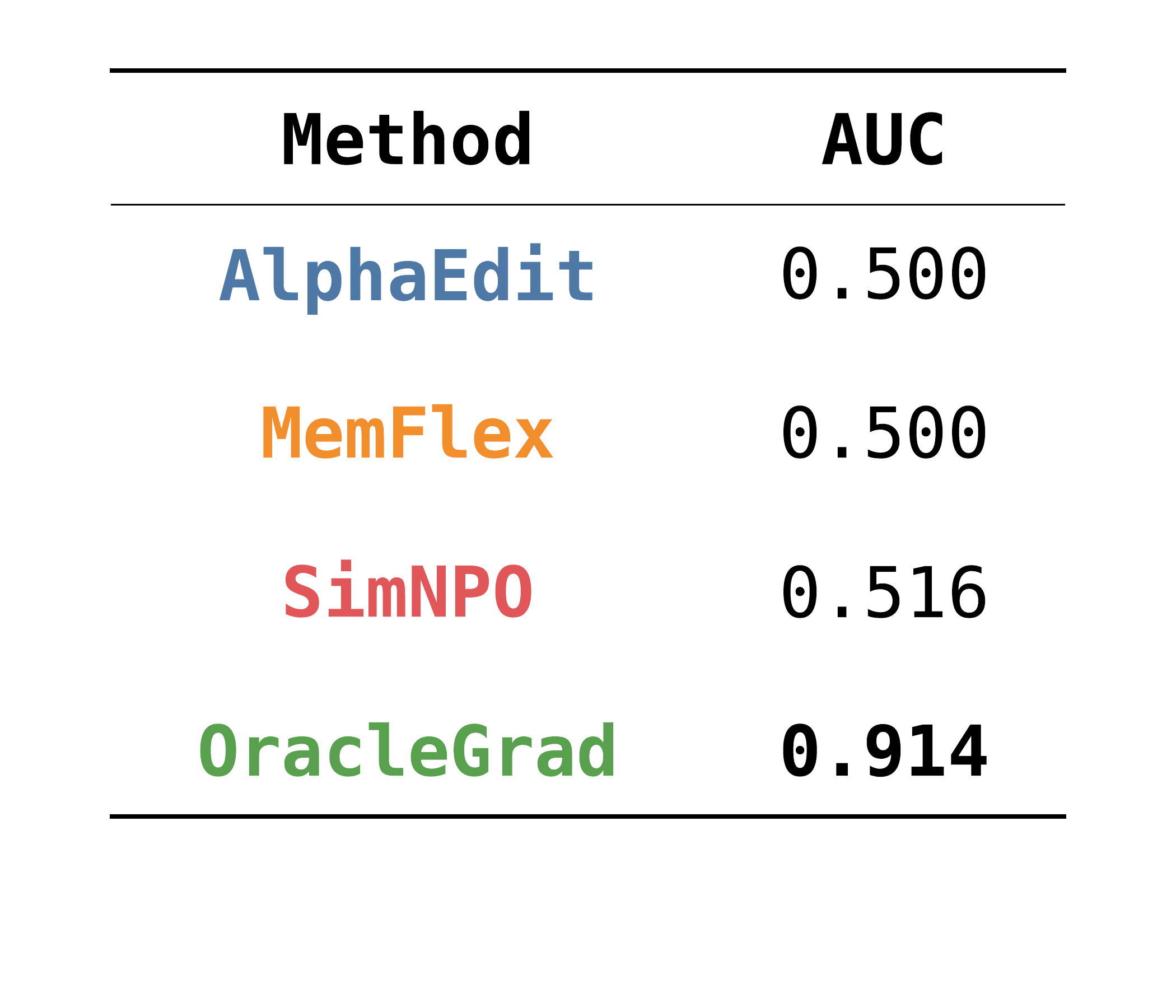}
    \caption{Localization precision,\\\texttt{driver's license}}
    \label{fig:localization-precision-1B-driverslicense}
  \end{subfigure}
  \begin{subfigure}[b]{0.72\textwidth}
    \includegraphics[width=\textwidth]{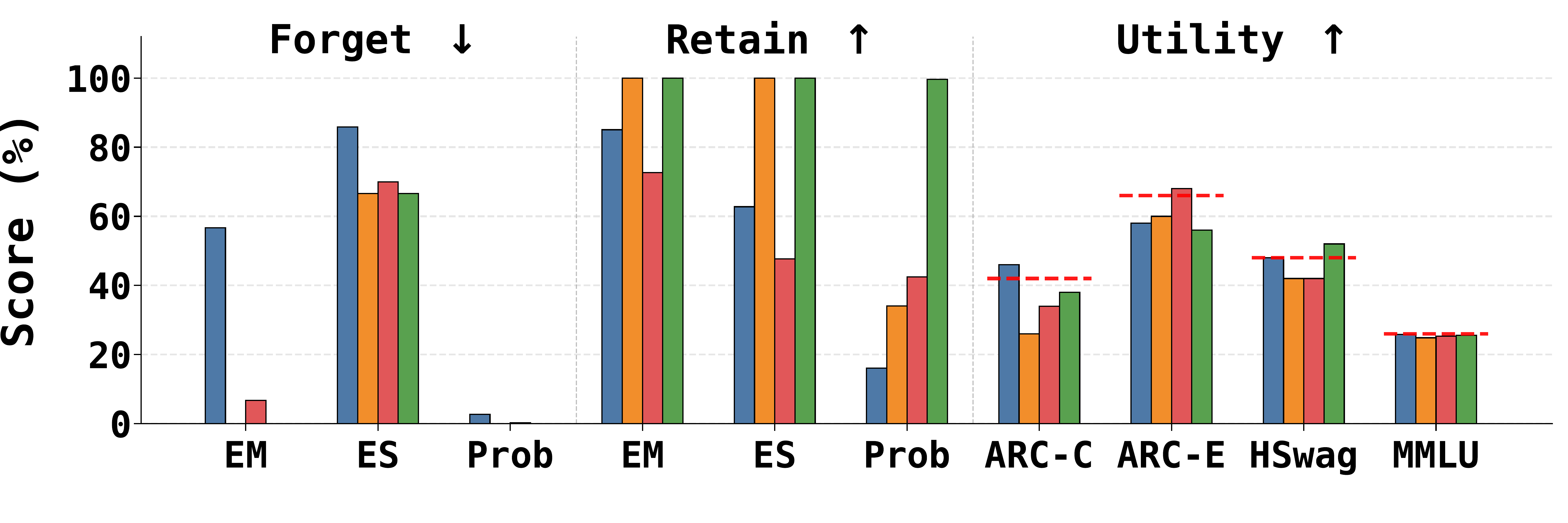}%
    \caption{Output-level evaluation,\\\texttt{birth city}}
    \label{fig:output-level-eval-1B-birthcity}
  \end{subfigure}%
  \begin{subfigure}[b]{0.28\textwidth}
    \includegraphics[width=\textwidth]{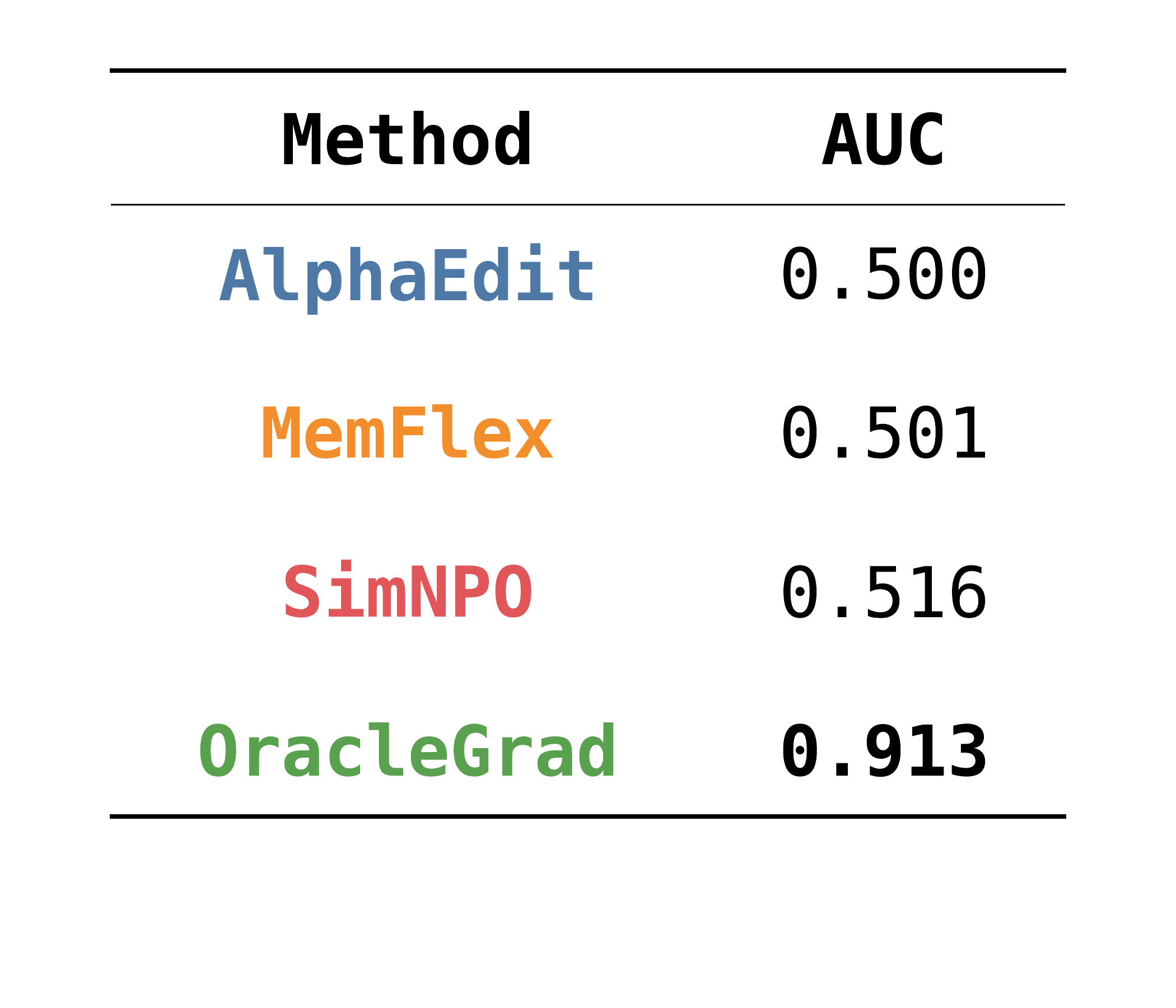}
    \caption{Localization precision,\\\texttt{birth city}}
    \label{fig:localization-precision-1B-birthcity}
  \end{subfigure}
  \begin{subfigure}[b]{0.72\textwidth}
    \includegraphics[width=\textwidth]{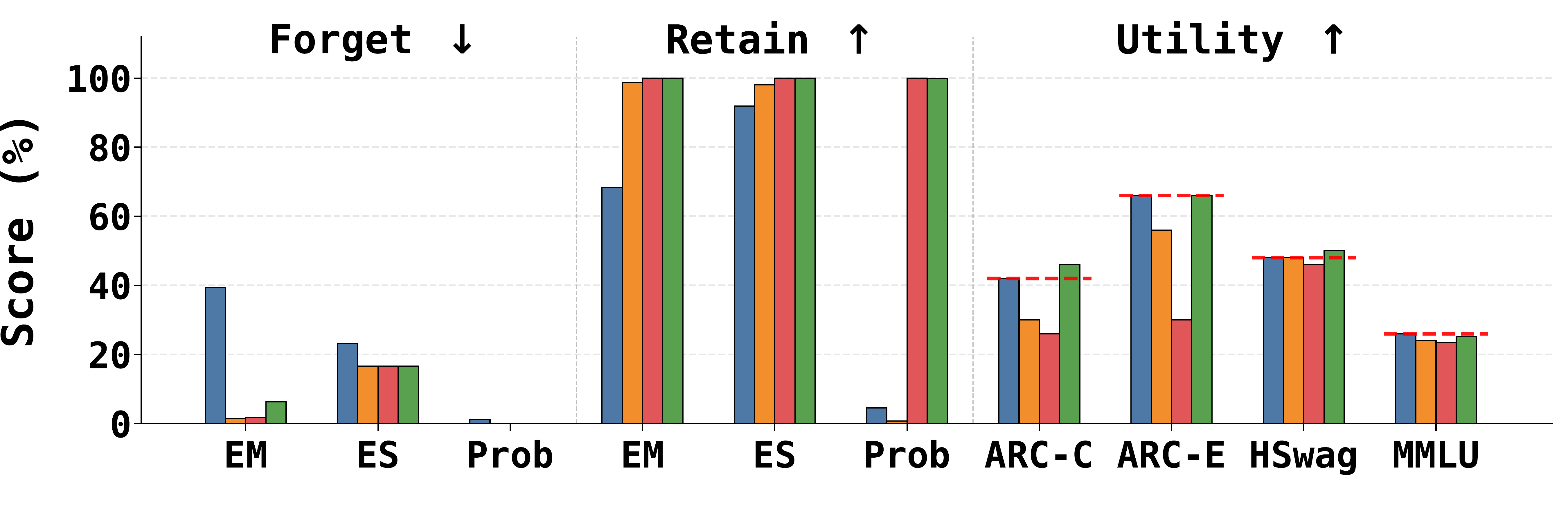}%
    \caption{Output-level evaluation,\\\texttt{phone number}}
    \label{fig:output-level-eval-1B-phonenumber}
  \end{subfigure}%
  \begin{subfigure}[b]{0.28\textwidth}
    \includegraphics[width=\textwidth]{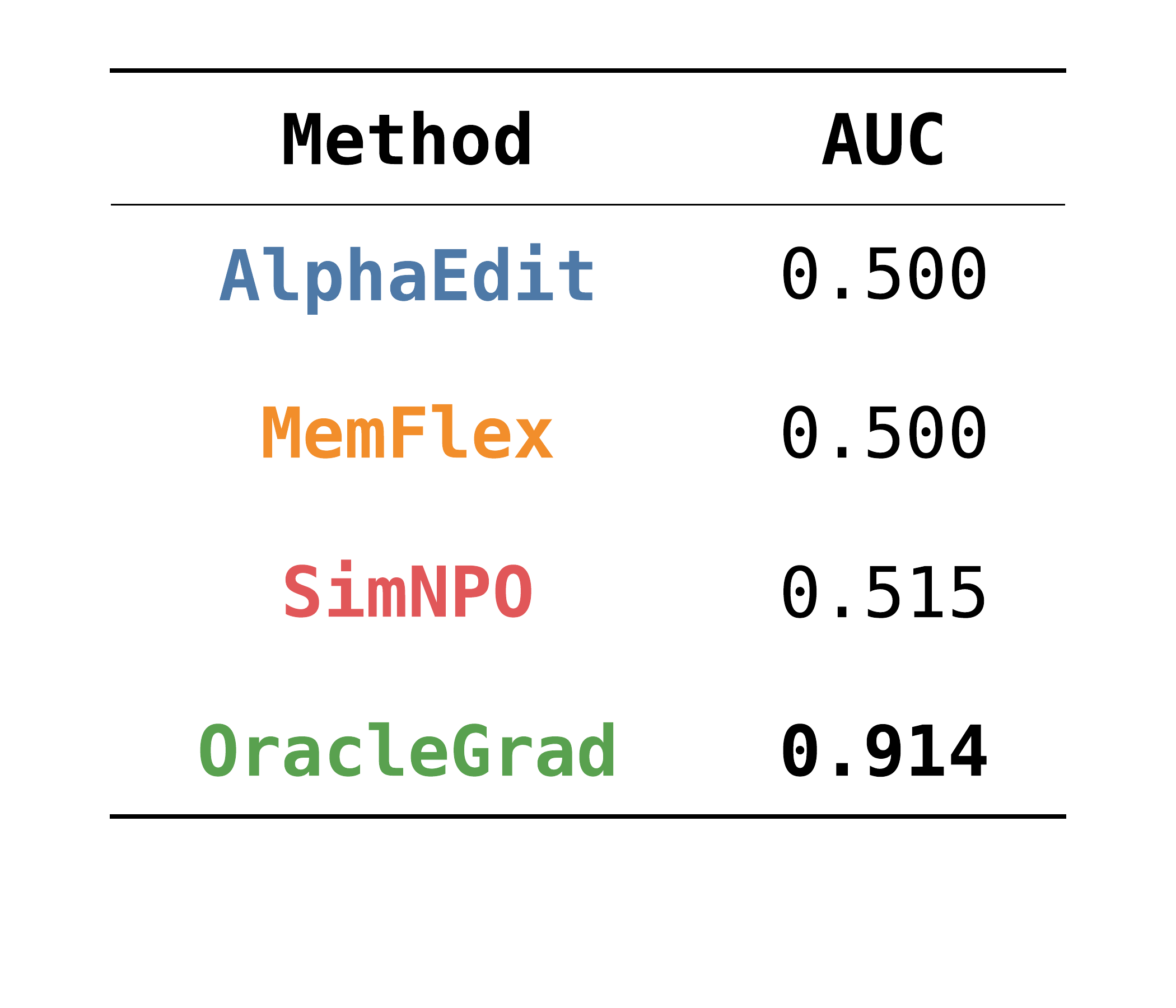}
    \caption{Localization precision,\\\texttt{phone number}}
    \label{fig:localization-precision-1B-phonenumber}
  \end{subfigure}
  
  \caption{Unlearning Evaluation for \olmosmall{}, for three PII fields (see the main paper for \emailaddress). On the left-hand side, we visualize Forget, Retain, and Utility metrics. The \protect\tikz[baseline=-0.5ex]{\protect\draw[red, dashed, line width=1pt] (0,0) -- (0.5,0);} on Utility represents the {\color{red}\textbf{Pre-Unlearning}} results. On the right, we report the evaluation of our proposed \textbf{Localization Precision} metric over
  \alphaedit, \memflex, \simnpo, and our Oracle-baseline \oraclegrad.
    }
  \label{fig:unlearning_all_1B}
\end{figure}

\begin{figure}[!h]
  \centering
  \begin{subfigure}[b]{0.72\textwidth}
    \includegraphics[width=\textwidth]{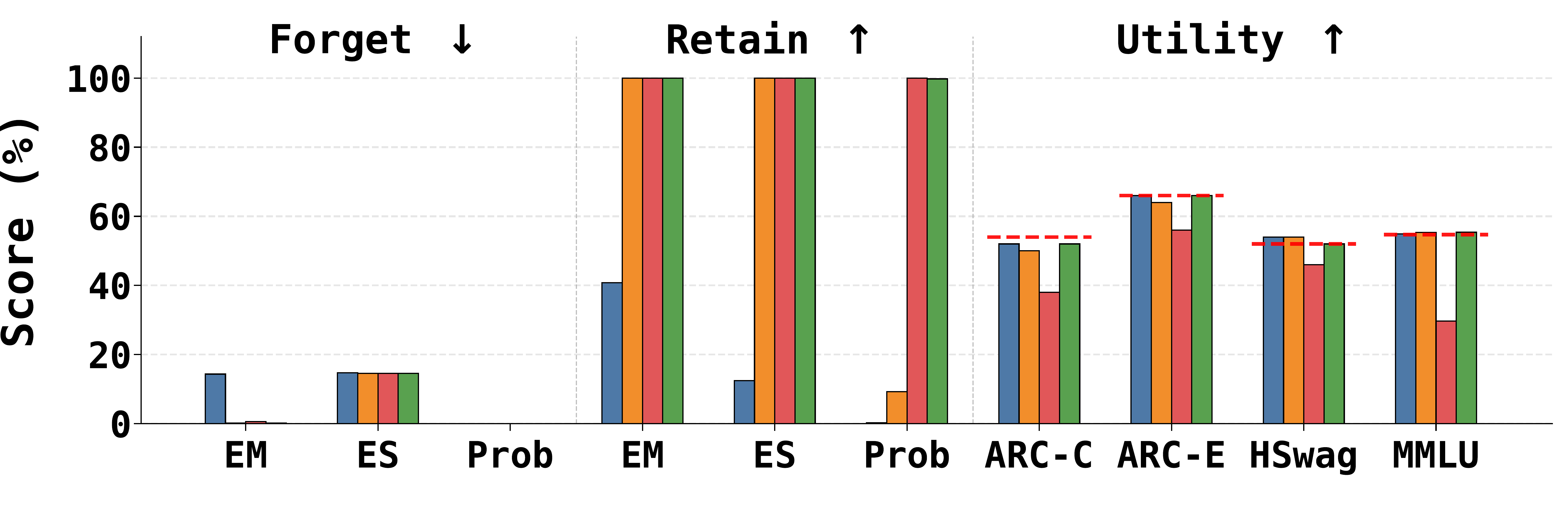}%
    \caption{Output-level evaluation,\\\texttt{driver's license}}
    \label{fig:output-level-eval-7B-driverslicense}
  \end{subfigure}%
  \begin{subfigure}[b]{0.28\textwidth}
    \includegraphics[width=\textwidth]{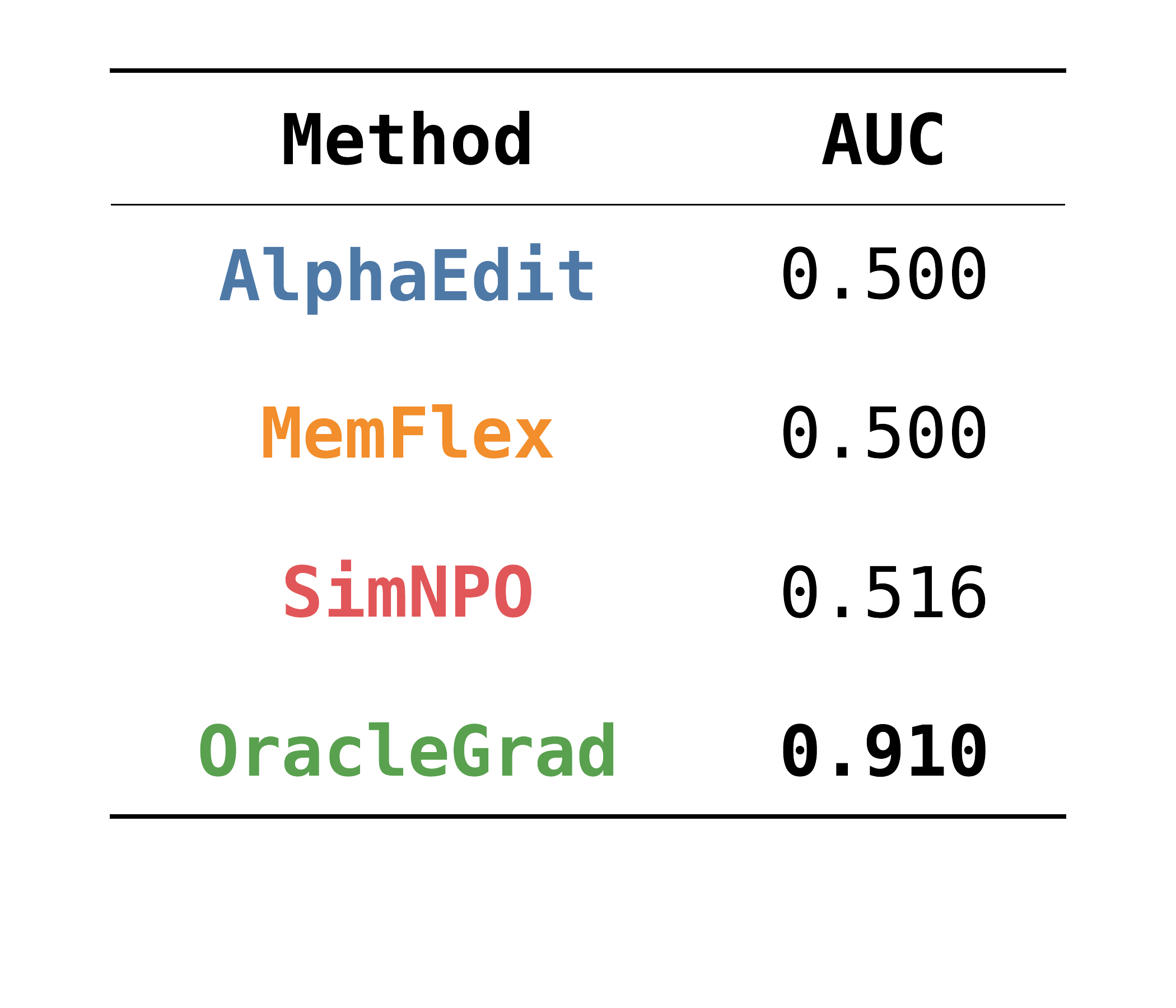}
    \caption{Localization precision,\\\texttt{driver's license}}
    \label{fig:localization-precision-7B-driverslicense}
  \end{subfigure}
  \begin{subfigure}[b]{0.72\textwidth}
    \includegraphics[width=\textwidth]{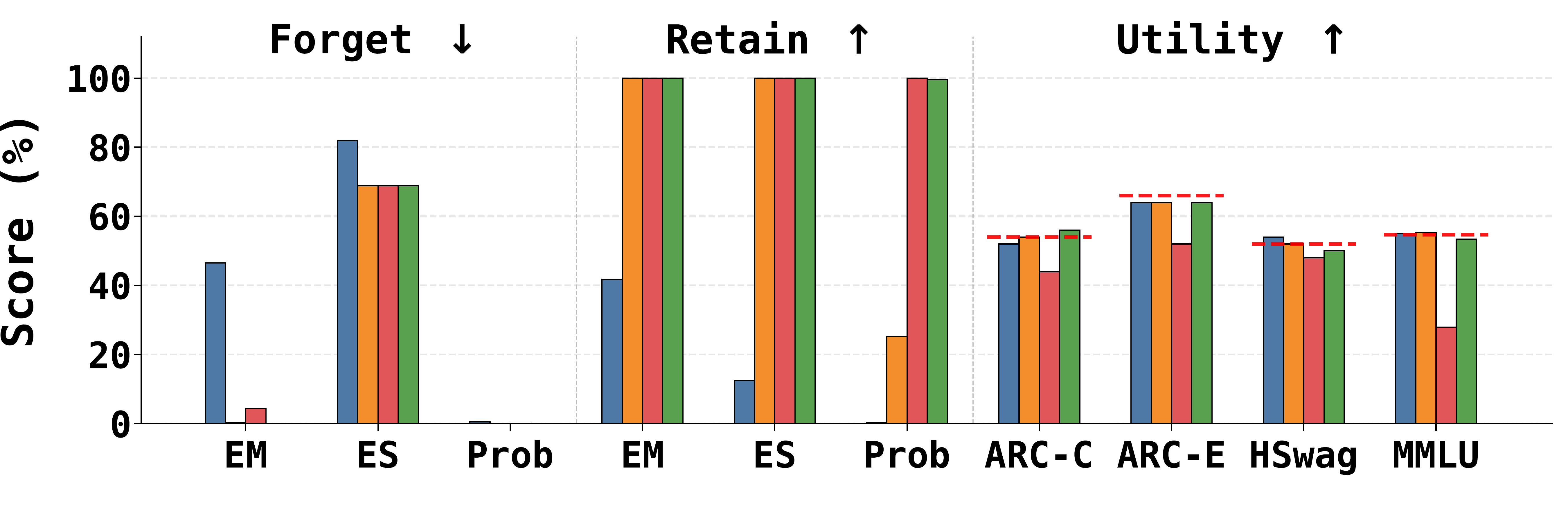}%
    \caption{Output-level evaluation,\\\texttt{birth city}}
    \label{fig:output-level-eval-7B-birthcity}
  \end{subfigure}%
  \begin{subfigure}[b]{0.28\textwidth}
    \includegraphics[width=\textwidth]{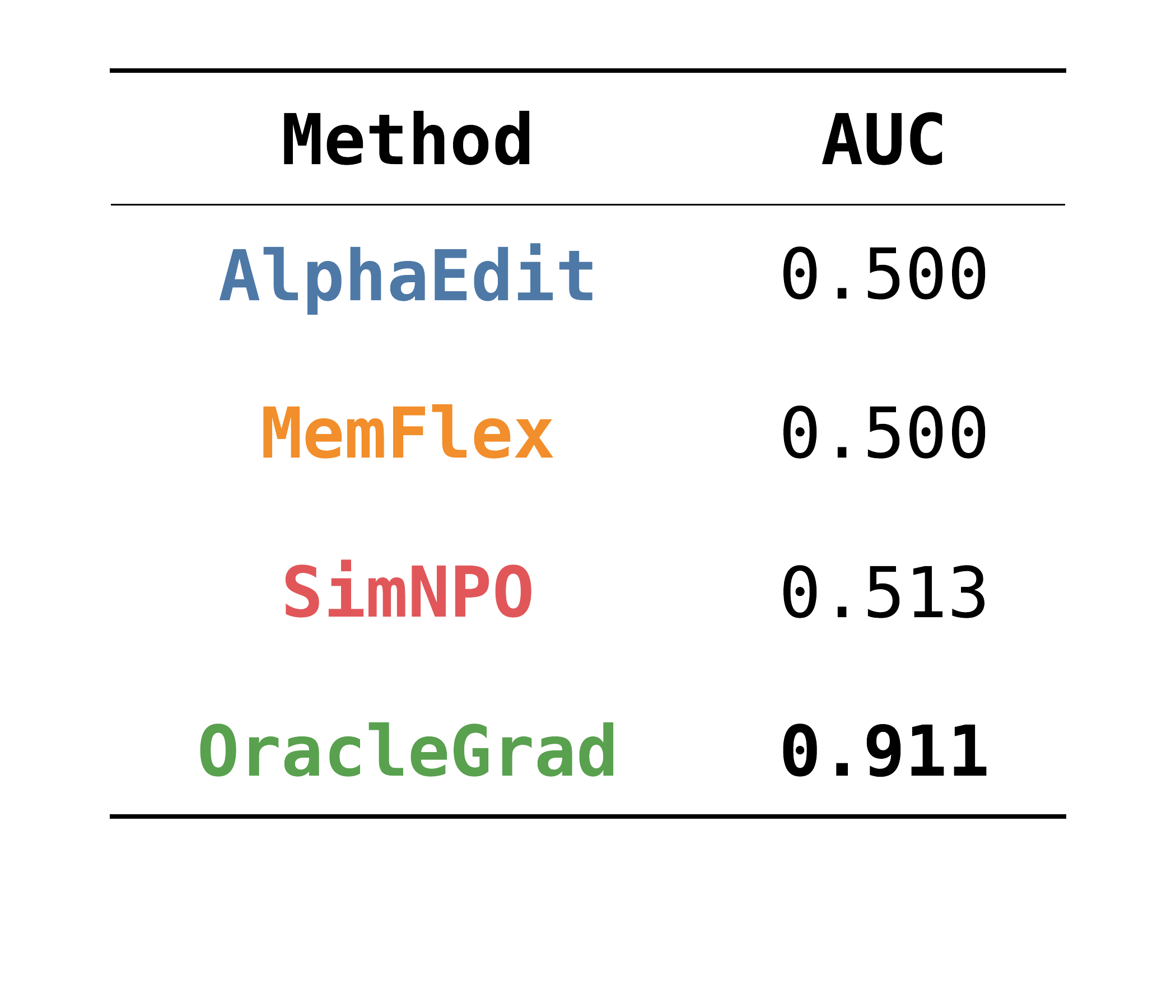}
    \caption{Localization precision,\\\texttt{birth city}}
    \label{fig:localization-precision-7B-birthcity}
  \end{subfigure}
  \begin{subfigure}[b]{0.72\textwidth}
    \includegraphics[width=\textwidth]{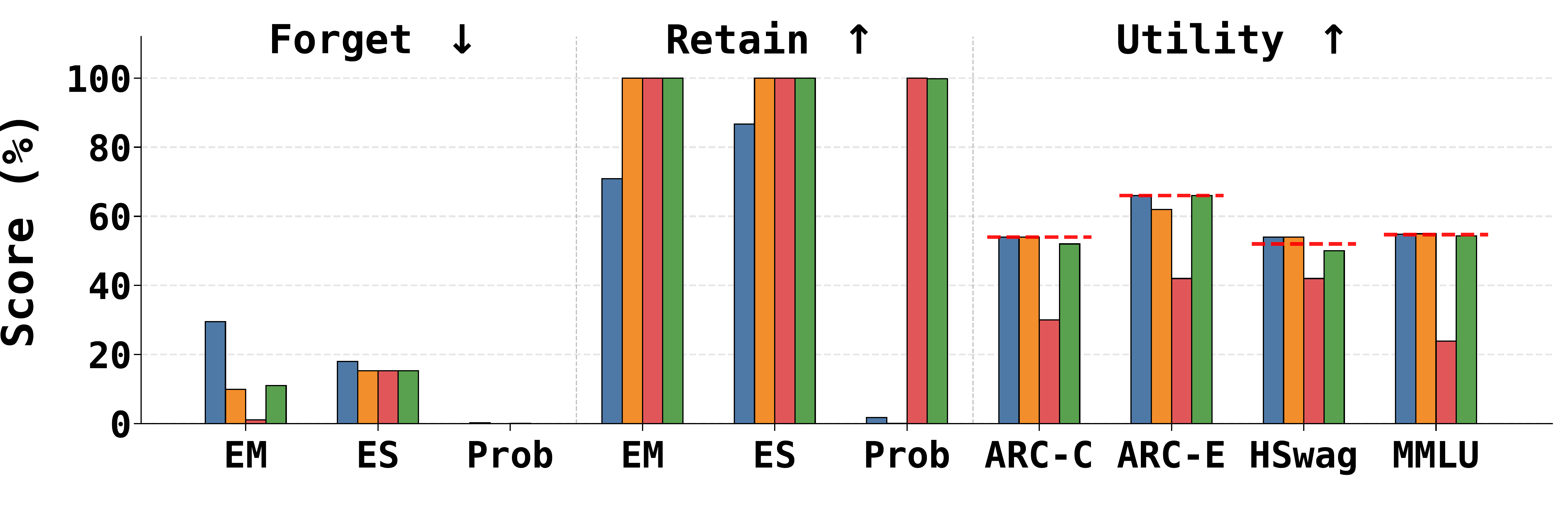}%
    \caption{Output-level evaluation,\\\texttt{phone number}}
    \label{fig:output-level-eval-7B-phonenumber}
  \end{subfigure}%
  \begin{subfigure}[b]{0.28\textwidth}
    \includegraphics[width=\textwidth]{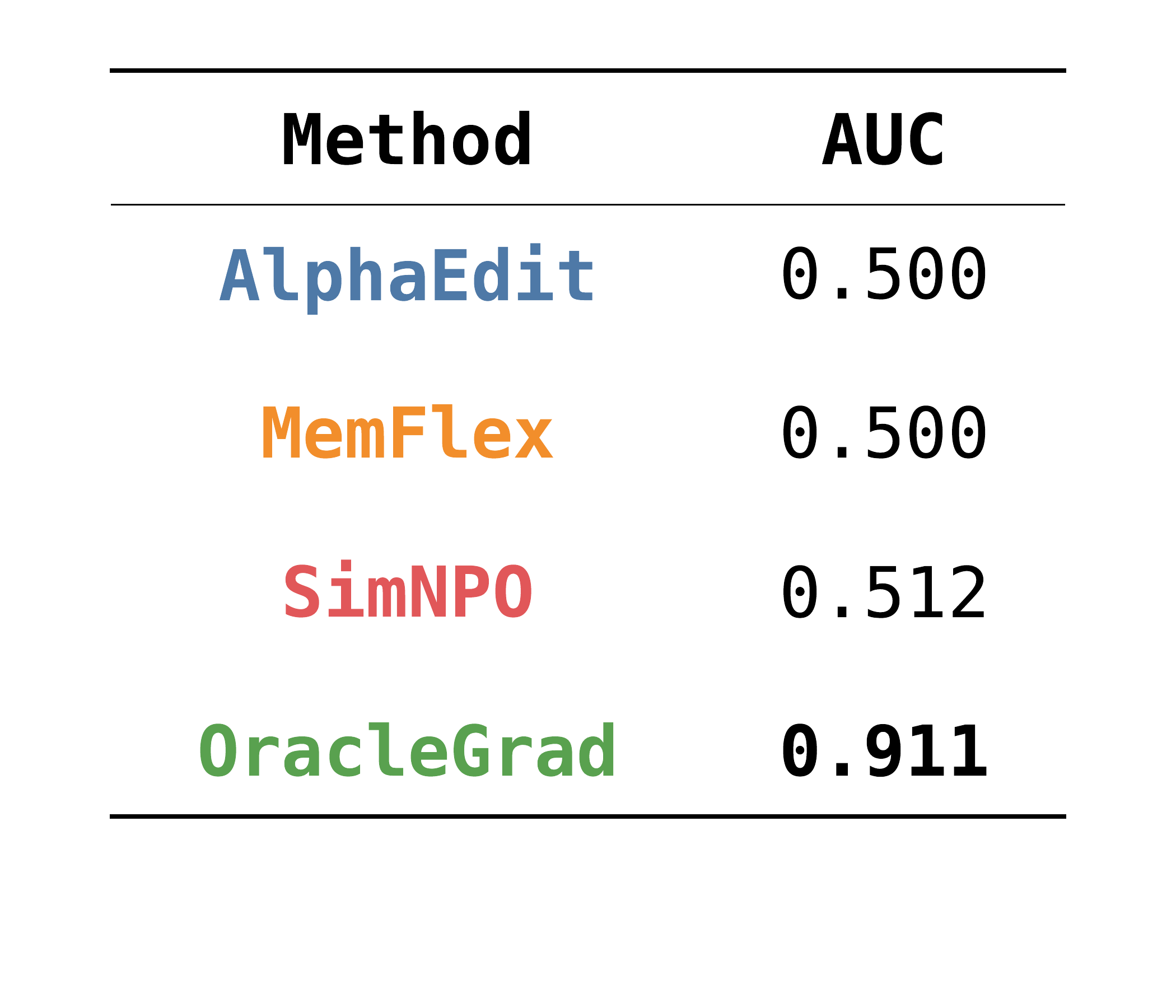}
    \caption{Localization precision,\\\texttt{phone number}}
    \label{fig:localization-precision-7B-phonenumber}
  \end{subfigure}
  \begin{subfigure}[b]{0.72\textwidth}
    \includegraphics[width=\textwidth]{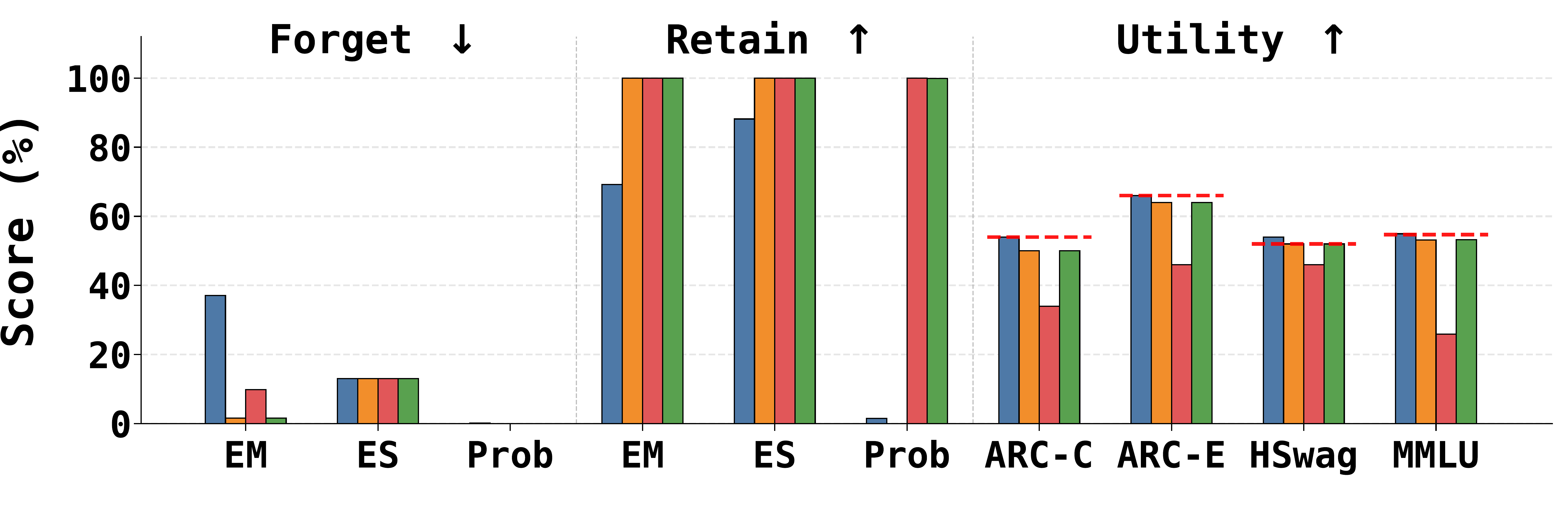}%
    \caption{Output-level evaluation,\\\texttt{email address}}
    \label{fig:output-level-eval-7B-emailaddress}
  \end{subfigure}%
  \begin{subfigure}[b]{0.28\textwidth}
    \includegraphics[width=\textwidth]{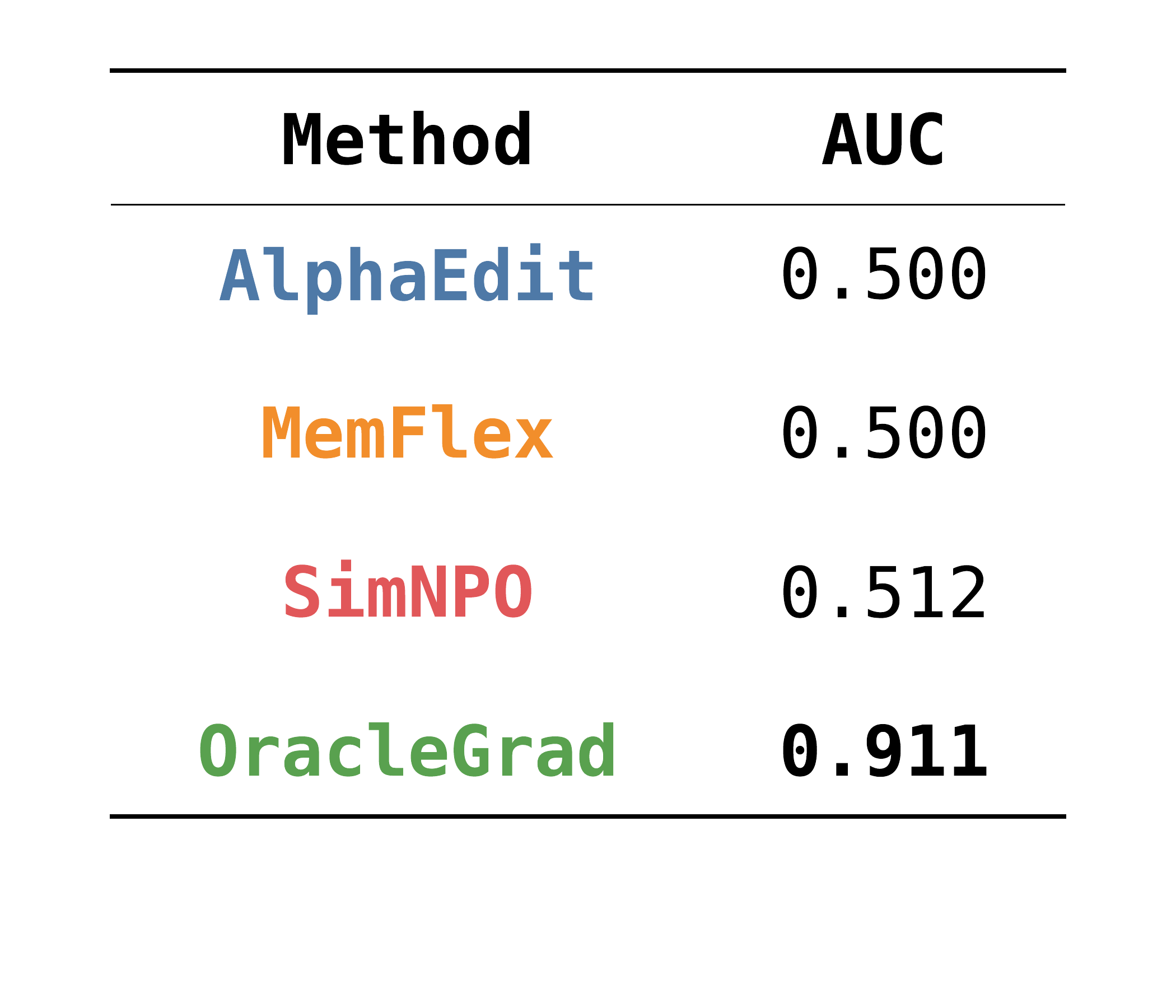}
    \caption{Localization precision,\\\texttt{email address}}
    \label{fig:localization-precision-7B-emailaddress}
  \end{subfigure}
  \captionof{figure}{Unlearning evaluation for \olmobig{}, for all four PII fields. On the left-hand side, we plot the Forget, Retain, and Utility metrics. The \protect\tikz[baseline=-0.5ex]{\protect\draw[red, dashed, line width=1pt] (0,0) -- (0.5,0);} on Utility represents the {\color{red}\textbf{Pre-Unlearning}} results. On the right, we report the evaluation of our proposed \textbf{Localization Precision} metric over
  \alphaedit, \memflex, \simnpo, and our Oracle-baseline \oraclegrad.
}
  \label{fig:unlearning_all_7B}
\end{figure}

In the main paper (\Cref{sec:unlearning_results}), we only included the unlearning evaluations for \texttt{email addresses} for the \olmosmall{} model, due to space constraints. Here, \Cref{fig:unlearning_all_1B} and \Cref{fig:unlearning_all_7B} include all remaining results.

\clearpage
\clearpage
Finally, for completeness, we present the results for all PII fields (as previously included in the figures) in Tables~\ref{tab:master_all_fields-1B} and~\ref{tab:master_all_fields-7B}, for the 1B and 7B models, respectively.

\begin{minipage}{\textwidth}
\centering\footnotesize
\captionof{table}{Unlearning Results - \olmosmall - Cumulative results for all unlearning methods: \textbf{AlphaEdit} (AE), \textbf{MemFlex} (MF), \textbf{OracleGrad} (OG), \textbf{SimNPO} (SN). AUC (F) uses the forget mask $m^F$ as in-mask ground truth, while AUC (F$|$R) uses the unified label $m^F \vee m^R$; the unified variant is undefined for \oraclegrad{} (which targets only $m^F$), shown as ---.}
\label{tab:master_all_fields-1B}
\begin{adjustbox}{max width=\textwidth}
\setlength{\tabcolsep}{3pt}
\begin{tabular}{ll|rrrr|rrrr|rrrr|rrrr}
\toprule
 &  & \multicolumn{4}{c}{Email Address} & \multicolumn{4}{c}{Phone Number} & \multicolumn{4}{c}{Birth City} & \multicolumn{4}{c}{Driver's License} \\
 &  & AE & MF & OG & SN & AE & MF & OG & SN & AE & MF & OG & SN & AE & MF & OG & SN \\
\midrule
\multirow{6}{*}{Forget} & EM & 63.2 & 36.8 & \textbf{1.6} & 17.7 & 39.4 & \textbf{1.5} & 6.3 & 1.7 & 56.6 & \textbf{0.0} & \textbf{0.0} & 6.8 & 64.9 & \textbf{0.0} & 0.1 & 0.9 \\
 & ES & 34.8 & \textbf{15.9} & \textbf{15.9} & 16.7 & 23.2 & \textbf{16.6} & \textbf{16.6} & \textbf{16.6} & 85.8 & \textbf{66.6} & \textbf{66.6} & 70.0 & 23.0 & \textbf{15.6} & \textbf{15.6} & \textbf{15.6} \\
 & EM Paraph. & 63.6 & 36.7 & \textbf{1.6} & 18.0 & 39.2 & \textbf{1.5} & 6.3 & 1.5 & 55.4 & \textbf{0.0} & \textbf{0.0} & 5.8 & 64.1 & \textbf{0.0} & 0.1 & 0.5 \\
 & ES Paraph. & 34.1 & \textbf{15.9} & \textbf{15.9} & 16.6 & 23.1 & \textbf{16.6} & \textbf{16.6} & \textbf{16.6} & 85.9 & \textbf{66.6} & \textbf{66.6} & 70.0 & 23.9 & \textbf{15.6} & \textbf{15.6} & \textbf{15.6} \\
 & Prob & 10.3 & 0.0 & \textbf{0.0} & 0.9 & 1.2 & \textbf{0.0} & \textbf{0.0} & 0.0 & 2.7 & \textbf{0.0} & \textbf{0.0} & 0.2 & 6.1 & \textbf{0.0} & \textbf{0.0} & 0.0 \\
 & Prob Paraph. & 10.6 & 0.0 & \textbf{0.0} & 0.9 & 1.3 & \textbf{0.0} & \textbf{0.0} & 0.0 & 2.6 & \textbf{0.0} & \textbf{0.0} & 0.2 & 6.2 & \textbf{0.0} & \textbf{0.0} & 0.0 \\
\midrule
\multirow{6}{*}{Retain} & EM & 58.8 & 98.0 & \textbf{100.0} & \textbf{100.0} & 68.3 & 98.8 & \textbf{100.0} & \textbf{100.0} & 85.1 & \textbf{100.0} & \textbf{100.0} & 72.7 & 81.7 & 96.9 & \textbf{100.0} & \textbf{100.0} \\
 & ES & 89.8 & 96.8 & \textbf{100.0} & \textbf{100.0} & 91.9 & 98.1 & \textbf{100.0} & \textbf{100.0} & 62.8 & \textbf{100.0} & \textbf{100.0} & 47.7 & 59.5 & 90.2 & \textbf{100.0} & \textbf{100.0} \\
 & EM Paraph. & 60.2 & 96.3 & \textbf{100.0} & 93.7 & 67.0 & 98.6 & \textbf{100.0} & \textbf{100.0} & 85.7 & 99.7 & \textbf{100.0} & 73.1 & 82.1 & 93.3 & \textbf{99.6} & 98.2 \\
 & ES Paraph. & 90.8 & 95.8 & \textbf{100.0} & 95.7 & 92.7 & 97.6 & \textbf{100.0} & \textbf{100.0} & 65.5 & 98.3 & \textbf{100.0} & 49.3 & 59.8 & 80.3 & \textbf{98.4} & 91.9 \\
 & Prob & 4.5 & 1.7 & 99.9 & \textbf{100.0} & 4.6 & 0.8 & 99.8 & \textbf{100.0} & 16.1 & 34.1 & \textbf{99.7} & 42.5 & 17.6 & 22.5 & 99.8 & \textbf{100.0} \\
 & Prob Paraph. & 4.5 & 1.7 & \textbf{99.7} & 91.3 & 4.4 & 0.8 & \textbf{99.7} & 99.6 & 16.7 & 34.0 & \textbf{98.7} & 43.4 & 18.0 & 22.6 & \textbf{98.5} & 95.8 \\
\midrule
\multirow{4}{*}{Utility ($\Delta$)} & ARC-C & \textbf{+4.0} & -4.0 & -2.0 & -12.0 & +0.0 & -12.0 & \textbf{+4.0} & -16.0 & \textbf{+4.0} & -16.0 & -4.0 & -8.0 & \textbf{+4.0} & +2.0 & -10.0 & -8.0 \\
 & ARC-E & +0.0 & -6.0 & \textbf{+2.0} & -18.0 & \textbf{+0.0} & -10.0 & \textbf{+0.0} & -36.0 & -8.0 & -6.0 & -10.0 & \textbf{+2.0} & \textbf{-2.0} & -6.0 & -4.0 & -16.0 \\
 & HSwag & \textbf{+0.0} & \textbf{+0.0} & -4.0 & -4.0 & +0.0 & +0.0 & \textbf{+2.0} & -2.0 & +0.0 & -6.0 & \textbf{+4.0} & -6.0 & \textbf{+2.0} & -2.0 & -4.0 & -2.0 \\
 & MMLU & \textbf{+0.1} & -1.5 & -1.2 & -1.8 & \textbf{+0.0} & -1.9 & -0.9 & -2.6 & \textbf{-0.2} & -1.2 & -0.4 & -0.7 & \textbf{-0.4} & -1.1 & -1.1 & -2.9 \\
\midrule
\multirow{2}{*}{Precision} & AUC (F|R) & 0.500 & 0.500 & \multicolumn{1}{c}{---} & \textbf{0.519} & 0.500 & 0.500 & \multicolumn{1}{c}{---} & \textbf{0.520} & 0.500 & 0.501 & \multicolumn{1}{c}{---} & \textbf{0.522} & 0.500 & 0.500 & \multicolumn{1}{c}{---} & \textbf{0.522} \\
 & AUC (F) & 0.500 & 0.500 & \textbf{0.915} & 0.515 & 0.500 & 0.500 & \textbf{0.914} & 0.515 & 0.500 & 0.501 & \textbf{0.913} & 0.516 & 0.500 & 0.500 & \textbf{0.914} & 0.516 \\
\bottomrule
\end{tabular}
\end{adjustbox}
\end{minipage}

\begin{minipage}[t]{\textwidth}
\centering\footnotesize
\captionof{table}{Unlearning Results - \olmobig - Cumulative results for all unlearning methods: \textbf{AlphaEdit} (AE), \textbf{MemFlex} (MF), \textbf{OracleGrad} (OG), \textbf{SimNPO} (SN). AUC (F) uses the forget mask $m^F$ as in-mask ground truth, while AUC (F$|$R) uses the unified label $m^F \vee m^R$; the unified variant is undefined for \oraclegrad{} (which targets only $m^F$), shown as ---.}
\label{tab:master_all_fields-7B}
\begin{adjustbox}{max width=\textwidth}
\setlength{\tabcolsep}{3pt}
\begin{tabular}{ll|rrrr|rrrr|rrrr|rrrr}
\toprule
 &  & \multicolumn{4}{c}{Email} & \multicolumn{4}{c}{Phone} & \multicolumn{4}{c}{Birth City} & \multicolumn{4}{c}{Driver's Lic.} \\
 &  & AE & MF & OG & SN & AE & MF & OG & SN & AE & MF & OG & SN & AE & MF & OG & SN \\
\midrule
\multirow{6}{*}{Forget} & EM & 37.1 & \textbf{1.6} & \textbf{1.6} & 9.8 & 29.5 & 9.9 & 11.0 & \textbf{1.1} & 46.5 & 0.3 & \textbf{0.0} & 4.4 & 14.3 & \textbf{0.1} & \textbf{0.1} & 0.6 \\
 & ES & \textbf{13.1} & \textbf{13.1} & \textbf{13.1} & \textbf{13.1} & 18.0 & \textbf{15.3} & \textbf{15.3} & \textbf{15.3} & 82.0 & \textbf{68.9} & \textbf{68.9} & \textbf{68.9} & 14.7 & \textbf{14.5} & \textbf{14.5} & \textbf{14.5} \\
 & EM Paraph. & 36.3 & \textbf{1.6} & \textbf{1.6} & 10.3 & 29.7 & 9.9 & 11.0 & \textbf{1.1} & 47.1 & \textbf{0.0} & \textbf{0.0} & 6.5 & 14.6 & \textbf{0.1} & \textbf{0.1} & 0.7 \\
 & ES Paraph. & \textbf{13.1} & \textbf{13.1} & \textbf{13.1} & \textbf{13.1} & 17.9 & \textbf{15.3} & \textbf{15.3} & \textbf{15.3} & 81.2 & \textbf{68.9} & \textbf{68.9} & \textbf{68.9} & 14.9 & \textbf{14.5} & \textbf{14.5} & \textbf{14.5} \\
 & Prob & 0.2 & \textbf{0.0} & \textbf{0.0} & 0.0 & 0.2 & \textbf{0.0} & \textbf{0.0} & 0.1 & 0.5 & \textbf{0.0} & \textbf{0.0} & 0.0 & 0.0 & \textbf{0.0} & \textbf{0.0} & 0.0 \\
 & Prob Paraph. & 0.2 & \textbf{0.0} & \textbf{0.0} & 0.0 & 0.2 & \textbf{0.0} & \textbf{0.0} & 0.1 & 0.7 & \textbf{0.0} & \textbf{0.0} & 0.1 & 0.0 & \textbf{0.0} & \textbf{0.0} & 0.0 \\
\midrule
\multirow{6}{*}{Retain} & EM & 69.2 & \textbf{100.0} & \textbf{100.0} & \textbf{100.0} & 70.9 & \textbf{100.0} & \textbf{100.0} & \textbf{100.0} & 41.8 & \textbf{100.0} & \textbf{100.0} & \textbf{100.0} & 40.8 & \textbf{100.0} & \textbf{100.0} & \textbf{100.0} \\
 & ES & 88.2 & \textbf{100.0} & \textbf{100.0} & \textbf{100.0} & 86.7 & \textbf{100.0} & \textbf{100.0} & \textbf{100.0} & 12.5 & \textbf{100.0} & \textbf{100.0} & \textbf{100.0} & 12.5 & \textbf{100.0} & \textbf{100.0} & \textbf{100.0} \\
 & EM Paraph. & 67.6 & \textbf{100.0} & \textbf{100.0} & 99.2 & 69.3 & \textbf{100.0} & \textbf{100.0} & 99.0 & 41.1 & \textbf{99.5} & 99.4 & 96.5 & 40.6 & 99.4 & \textbf{99.8} & 96.0 \\
 & ES Paraph. & 87.6 & \textbf{100.0} & \textbf{100.0} & 99.2 & 87.6 & \textbf{100.0} & \textbf{100.0} & 99.0 & 12.5 & \textbf{97.3} & 97.2 & 81.3 & 12.5 & 96.9 & \textbf{99.4} & 82.1 \\
 & Prob & 1.5 & 0.0 & 99.9 & \textbf{100.0} & 1.7 & 0.0 & 99.8 & \textbf{100.0} & 0.2 & 25.2 & 99.6 & \textbf{100.0} & 0.2 & 9.2 & 99.8 & \textbf{100.0} \\
 & Prob Paraph. & 1.7 & 0.0 & \textbf{99.7} & 99.0 & 1.8 & 0.0 & \textbf{99.7} & 99.0 & 0.2 & 24.2 & \textbf{96.2} & 89.6 & 0.2 & 8.7 & \textbf{98.9} & 89.1 \\
\midrule
\multirow{4}{*}{Utility ($\Delta$)} & ARC-C & \textbf{+0.0} & -4.0 & -4.0 & -20.0 & \textbf{+0.0} & \textbf{+0.0} & -2.0 & -24.0 & -2.0 & +0.0 & \textbf{+2.0} & -10.0 & \textbf{-2.0} & -4.0 & \textbf{-2.0} & -16.0 \\
 & ARC-E & \textbf{+0.0} & -2.0 & -2.0 & -20.0 & \textbf{+0.0} & -4.0 & \textbf{+0.0} & -24.0 & \textbf{-2.0} & \textbf{-2.0} & \textbf{-2.0} & -14.0 & \textbf{+0.0} & -2.0 & \textbf{+0.0} & -10.0 \\
 & HSwag & \textbf{+2.0} & +0.0 & +0.0 & -6.0 & \textbf{+2.0} & \textbf{+2.0} & -2.0 & -10.0 & \textbf{+2.0} & +0.0 & -2.0 & -4.0 & \textbf{+2.0} & \textbf{+2.0} & +0.0 & -6.0 \\
 & MMLU & \textbf{+0.2} & -1.5 & -1.5 & -28.8 & +0.1 & \textbf{+0.2} & -0.4 & -30.8 & +0.4 & \textbf{+0.6} & -1.3 & -26.8 & +0.2 & +0.6 & \textbf{+0.7} & -25.1 \\
\midrule
\multirow{2}{*}{Precision} & AUC (F|R) & 0.500 & 0.500 & \multicolumn{1}{c}{---} & \textbf{0.515} & 0.500 & 0.500 & \multicolumn{1}{c}{---} & \textbf{0.514} & 0.500 & 0.500 & \multicolumn{1}{c}{---} & \textbf{0.516} & 0.500 & 0.500 & \multicolumn{1}{c}{---} & \textbf{0.520} \\
 & AUC (F) & 0.500 & 0.500 & \textbf{0.911} & 0.512 & 0.500 & 0.500 & \textbf{0.911} & 0.512 & 0.500 & 0.500 & \textbf{0.911} & 0.513 & 0.500 & 0.500 & \textbf{0.910} & 0.516 \\
\bottomrule
\end{tabular}
\end{adjustbox}
\end{minipage}

\subsection{Resurfacing results}
We report here the additional results for the relearning attacks we performed on the models.
We note that \simnpo is, among the compared methods, the strongest (together with our strong baseline \oraclegrad). Additionally, we observe that leaked profiles by \oraclegrad are almost exactly overlapped with the ones leaked by \simnpo, suggesting that these samples might simply be more challenging to forget in the first place, and hence easier to recover.
Across fields, we also observe that \olmobig is consistently much less prone to resurfacing than \olmosmall. We refrain from drawing strong conclusions from this, as it remains unclear whether it reflects a genuine property of the larger model or instead a limitation of our straightforward resurfacing attack at the 7B scale.
\label{appendix:resurfacing_results}

\begin{figure}[h]
  \centering
  \begin{subfigure}[t]{0.3\linewidth}
    \centering
    \includegraphics[width=\linewidth]{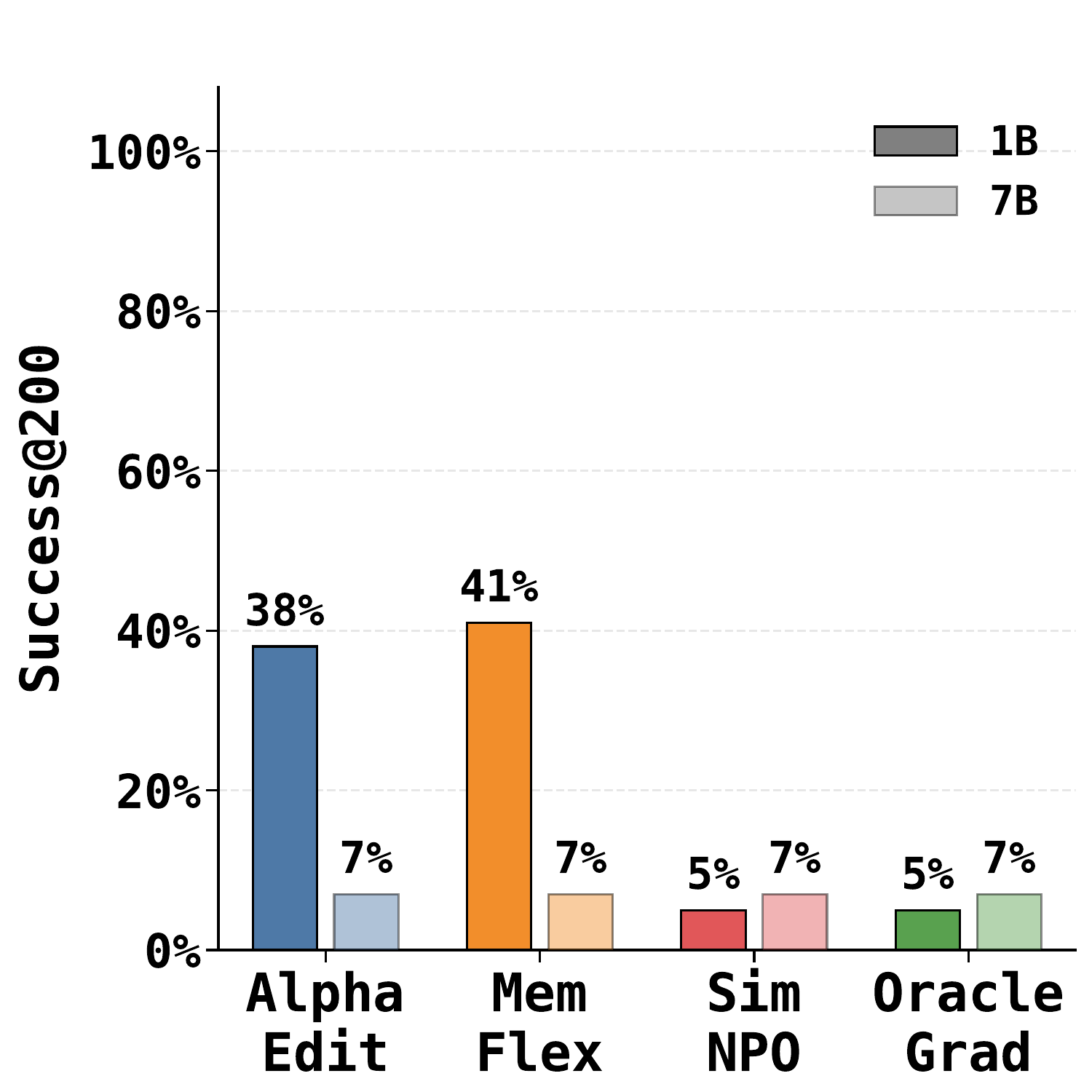}
    \caption{Relearning leakage rate.}
  \end{subfigure}
  \hfill
  \begin{subfigure}[t]{0.3\linewidth}
    \centering
    \includegraphics[width=\linewidth]{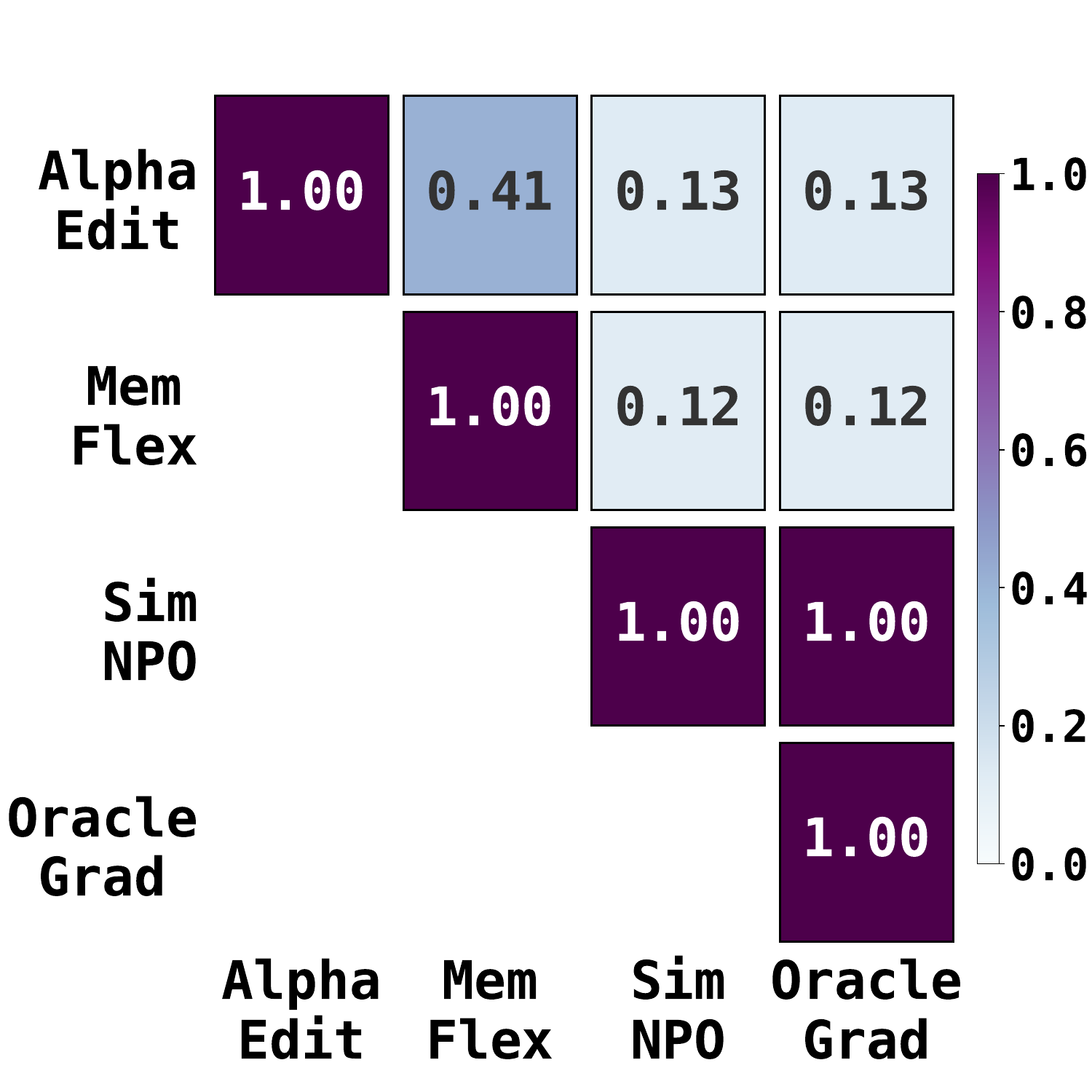}
    \caption{Leaked profiles Jaccard similarity - \olmosmall.}
  \end{subfigure}
  \hfill
  \begin{subfigure}[t]{0.3\linewidth}
    \centering
    \includegraphics[width=\linewidth]{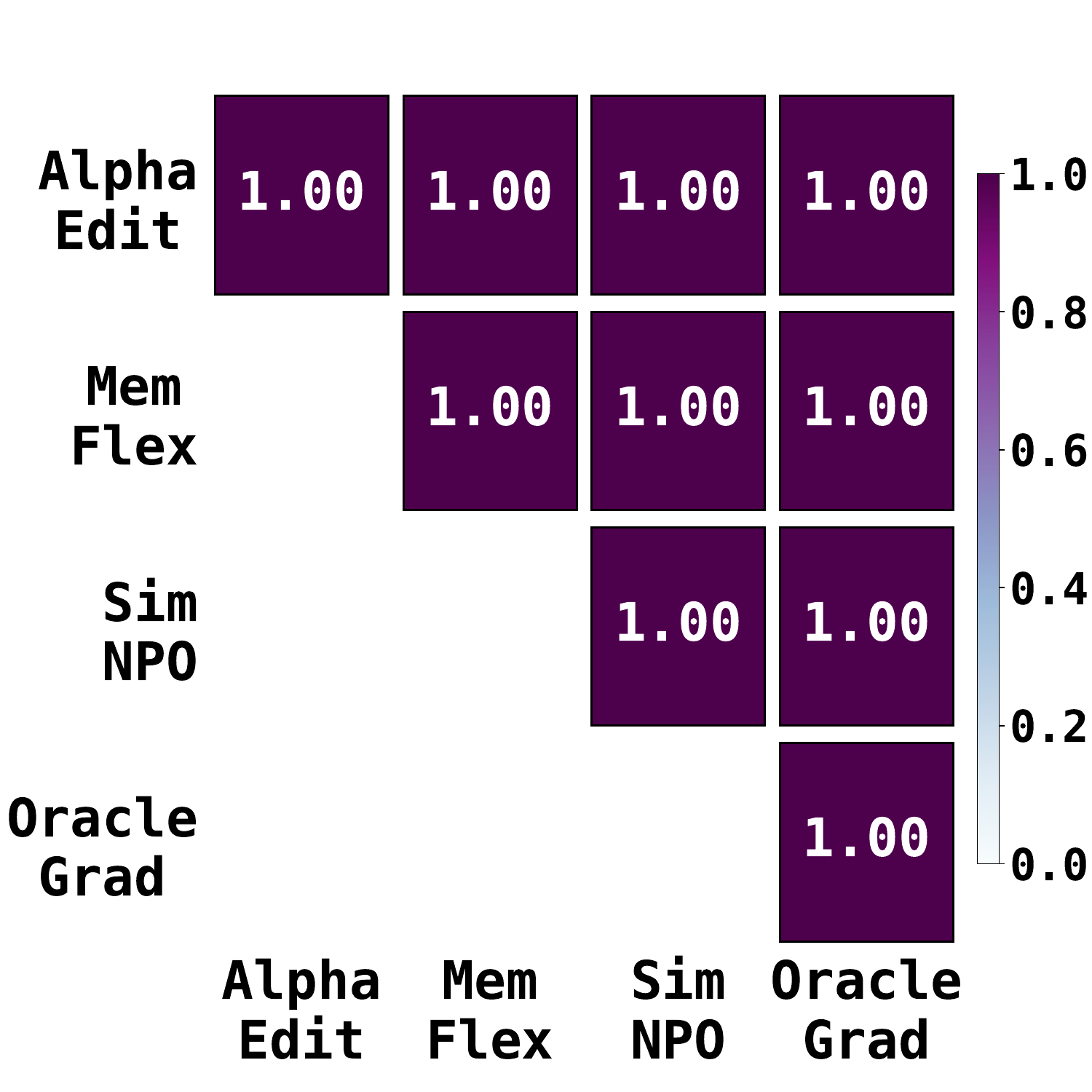}
    \caption{Leaked profiles Jaccard similarity - \olmobig.}
  \end{subfigure}
  \caption{Relearning vulnerability for \texttt{Driver's License}. The fact that \simnpo and \oraclegrad leak the same profile's information (Jaccard index of $1.00$) suggests that these profiles might be simply particularly challenging to forget.}
  \label{fig:relearn-drivers}
\end{figure}

\begin{figure}[h]
  \centering
  \begin{subfigure}[t]{0.3\linewidth}
    \centering
    \includegraphics[width=\linewidth]{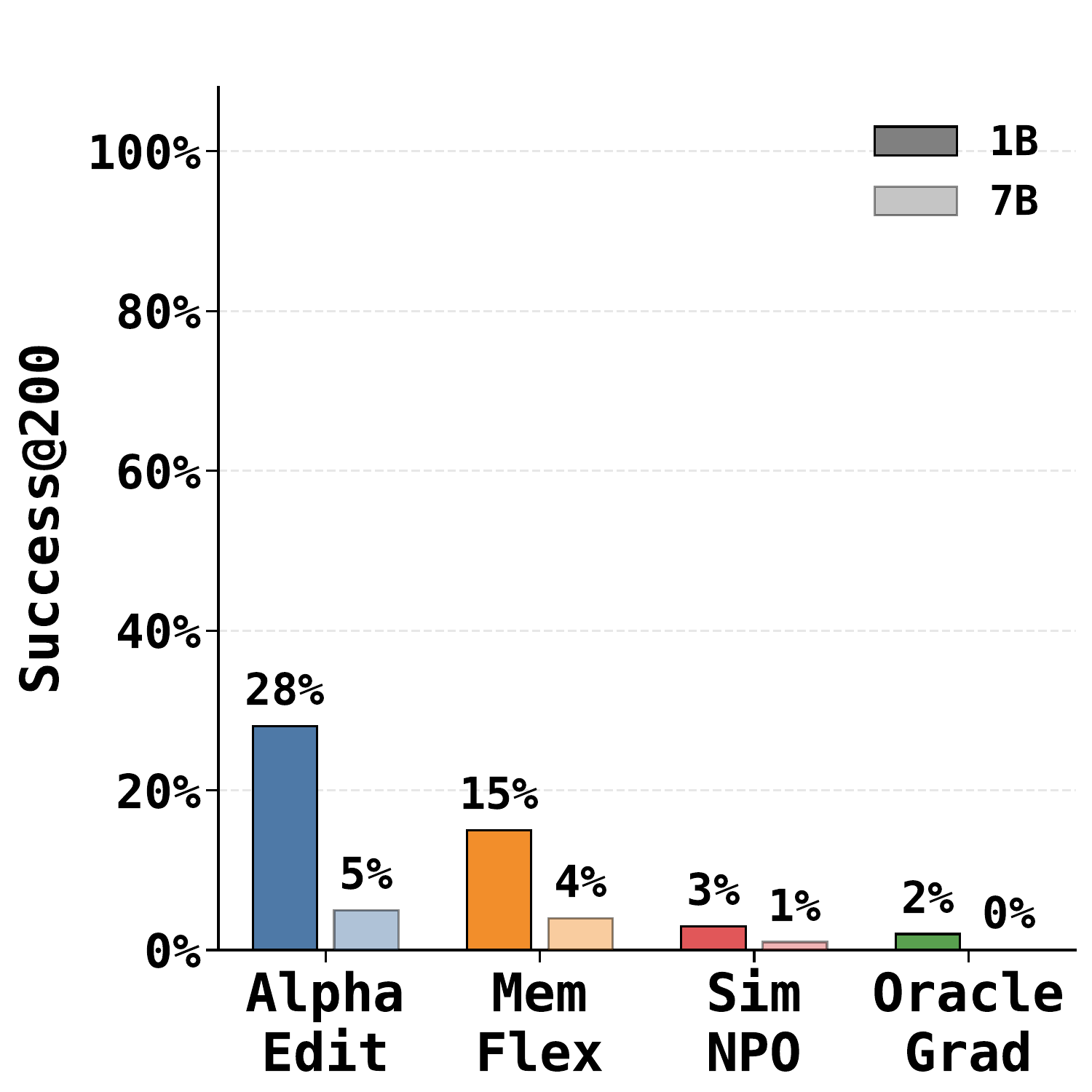}
    \caption{Relearning leakage rate.}
  \end{subfigure}
  \hfill
  \begin{subfigure}[t]{0.3\linewidth}
    \centering
    \includegraphics[width=\linewidth]{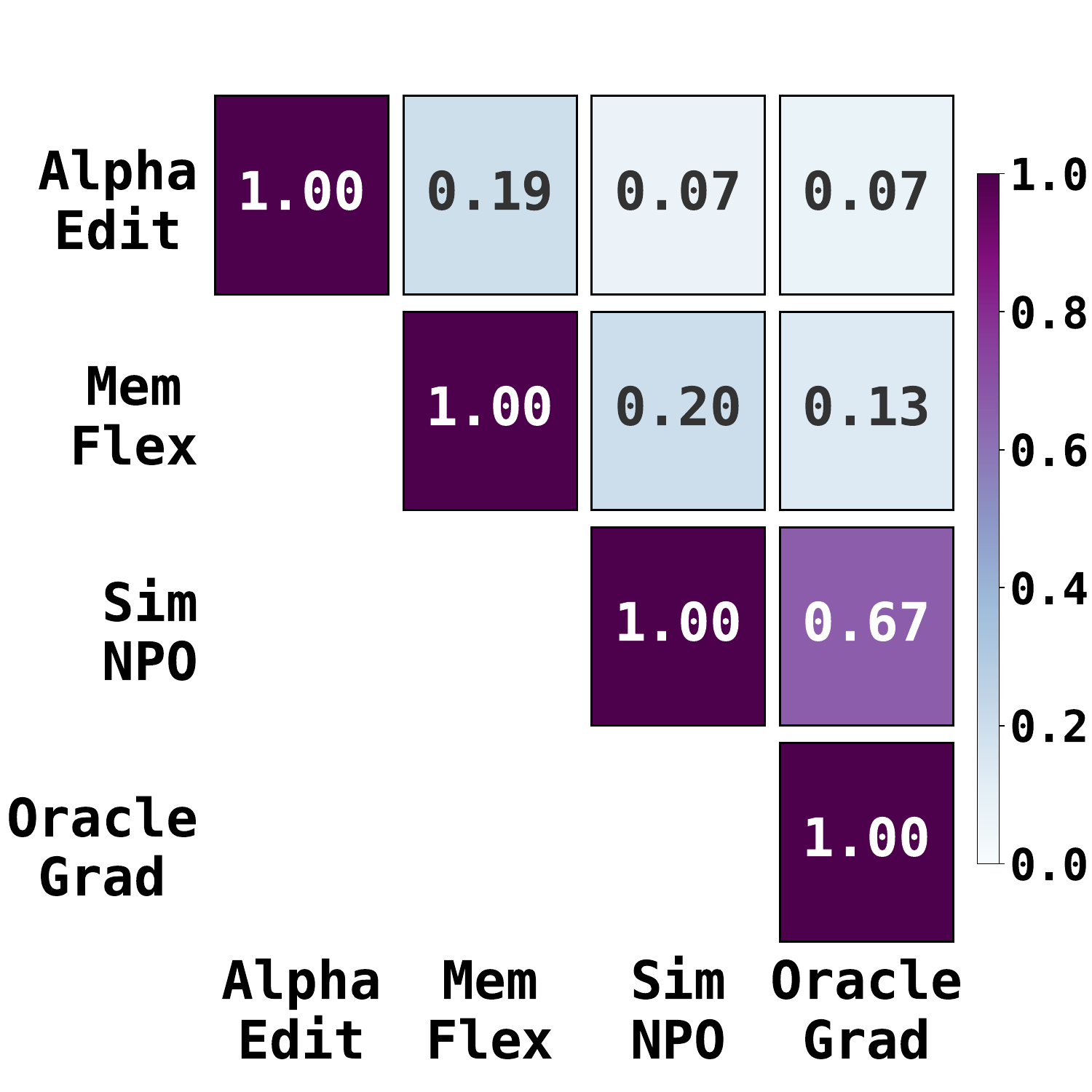}
    \caption{Leaked profiles Jaccard similarity - \olmosmall.}
  \end{subfigure}
  \hfill
  \begin{subfigure}[t]{0.3\linewidth}
    \centering
    \includegraphics[width=\linewidth]{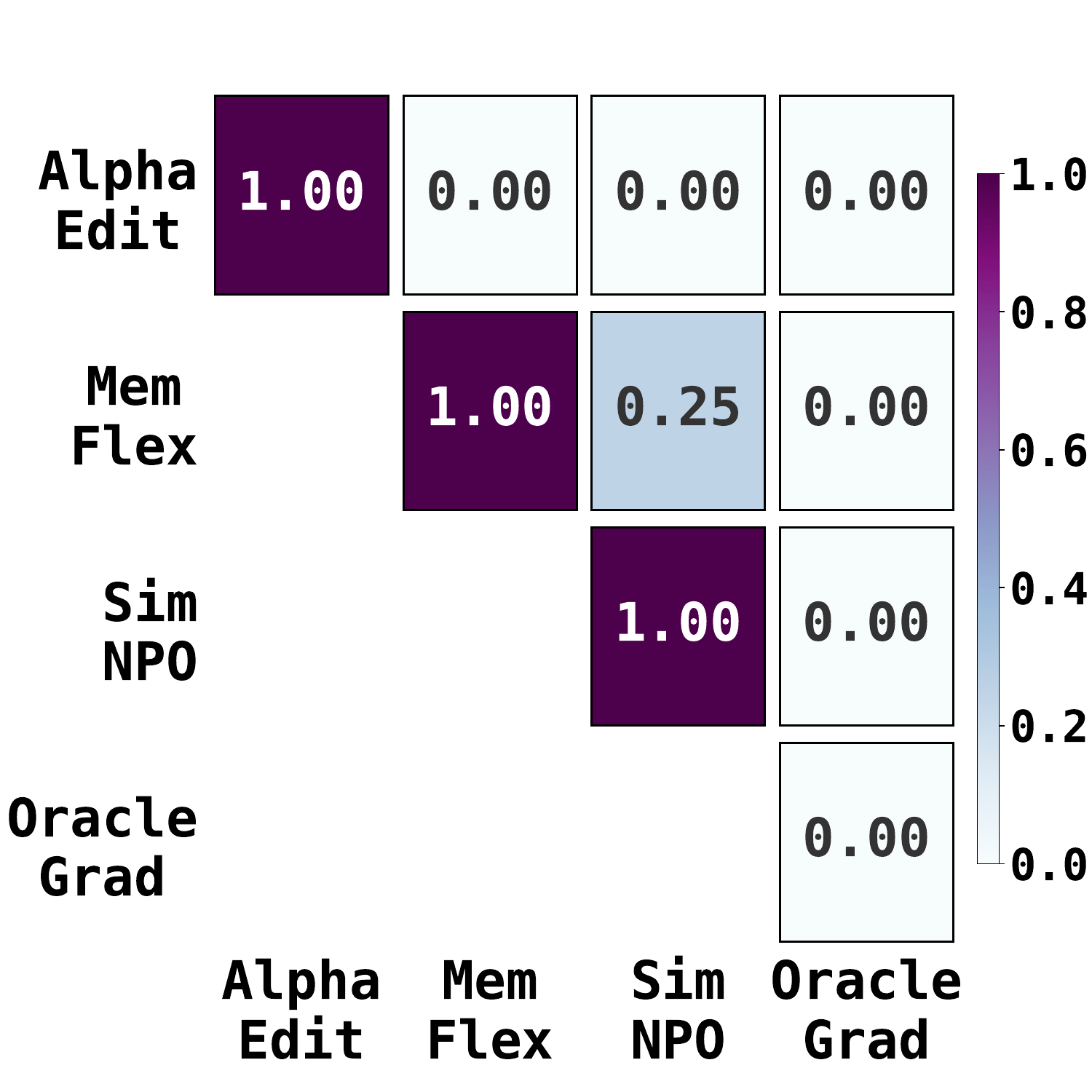}
    \caption{Leaked profiles Jaccard similarity - \olmobig.}
  \end{subfigure}
  \caption{Relearning vulnerability for \texttt{Birth City}.}
  \label{fig:relearn-birthcity}
\end{figure}

\begin{figure}[h]
  \centering
  \begin{subfigure}[t]{0.3\linewidth}
    \centering
    \includegraphics[width=\linewidth]{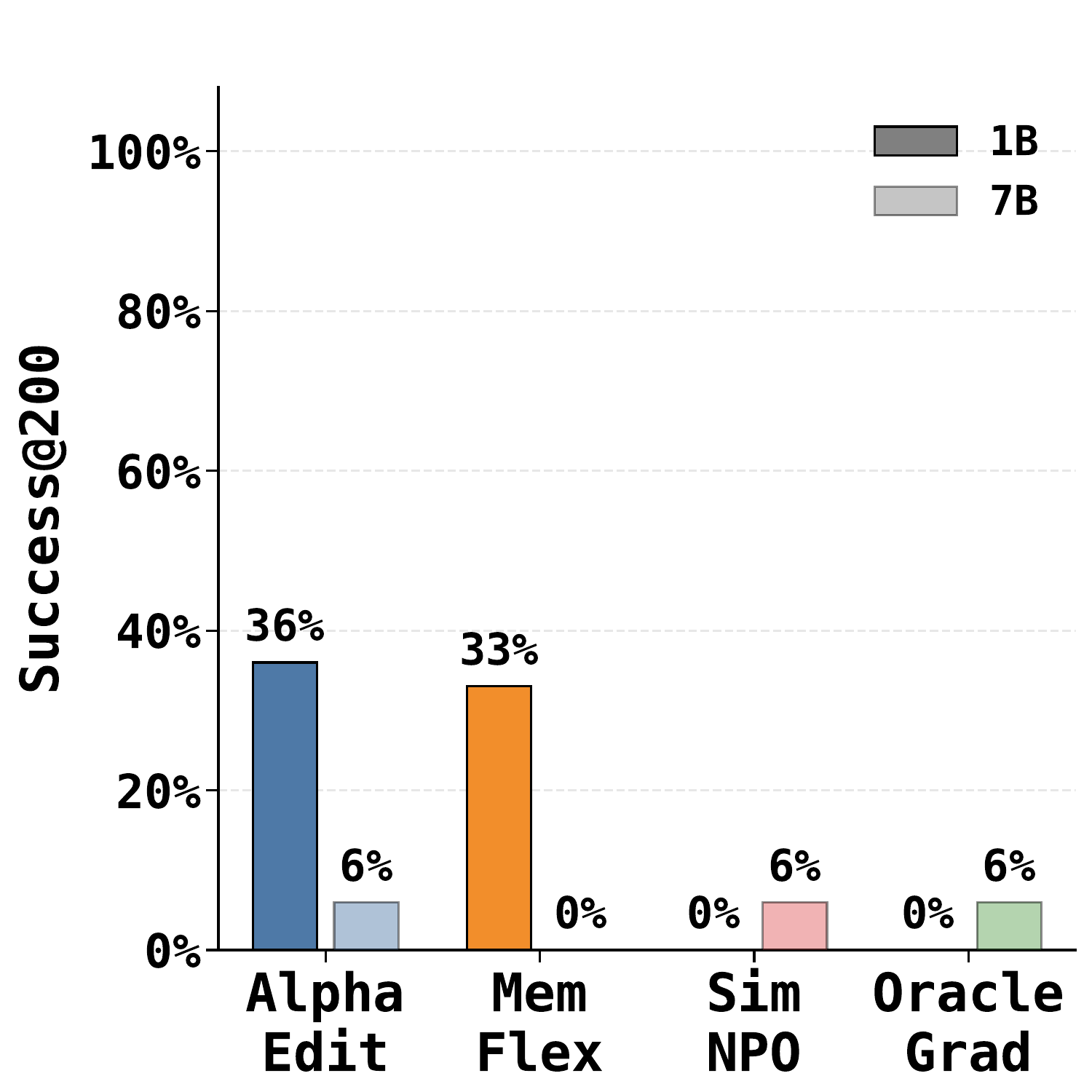}
    \caption{Relearning leakage rate.}
  \end{subfigure}
  \hfill
  \begin{subfigure}[t]{0.3\linewidth}
    \centering
    \includegraphics[width=\linewidth]{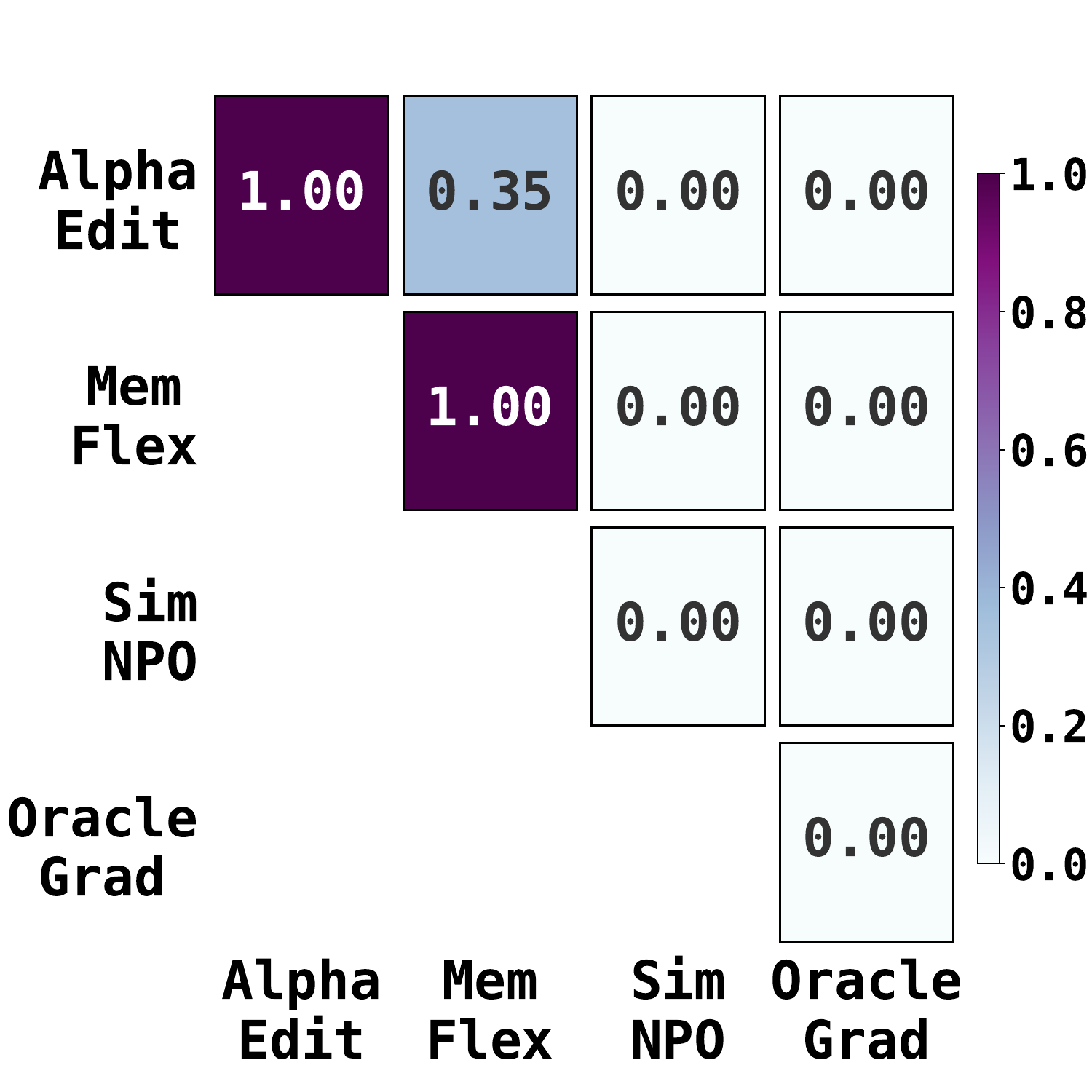}
    \caption{Leaked profiles Jaccard similarity - \olmosmall.}
  \end{subfigure}
  \hfill
  \begin{subfigure}[t]{0.3\linewidth}
    \centering
    \includegraphics[width=\linewidth]{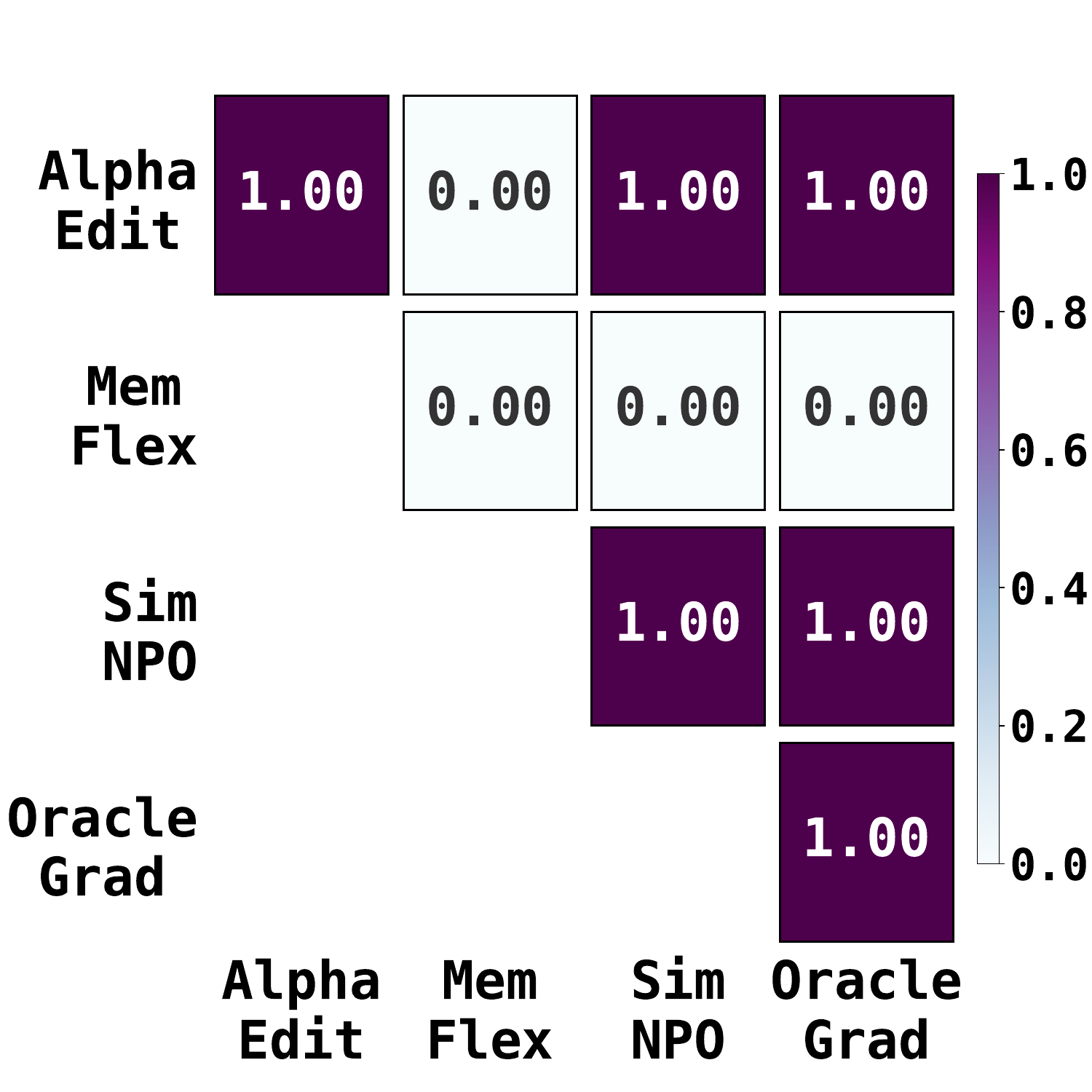}
    \caption{Leaked profiles Jaccard similarity - \olmobig.}
  \end{subfigure}
  \caption{Relearning vulnerability for \texttt{Phone Number}.}
  \label{fig:relearn-phonenumber}
\end{figure}

\begin{figure}[H]
  \centering
  \begin{subfigure}[t]{0.3\linewidth}
    \centering
    \includegraphics[width=\linewidth]{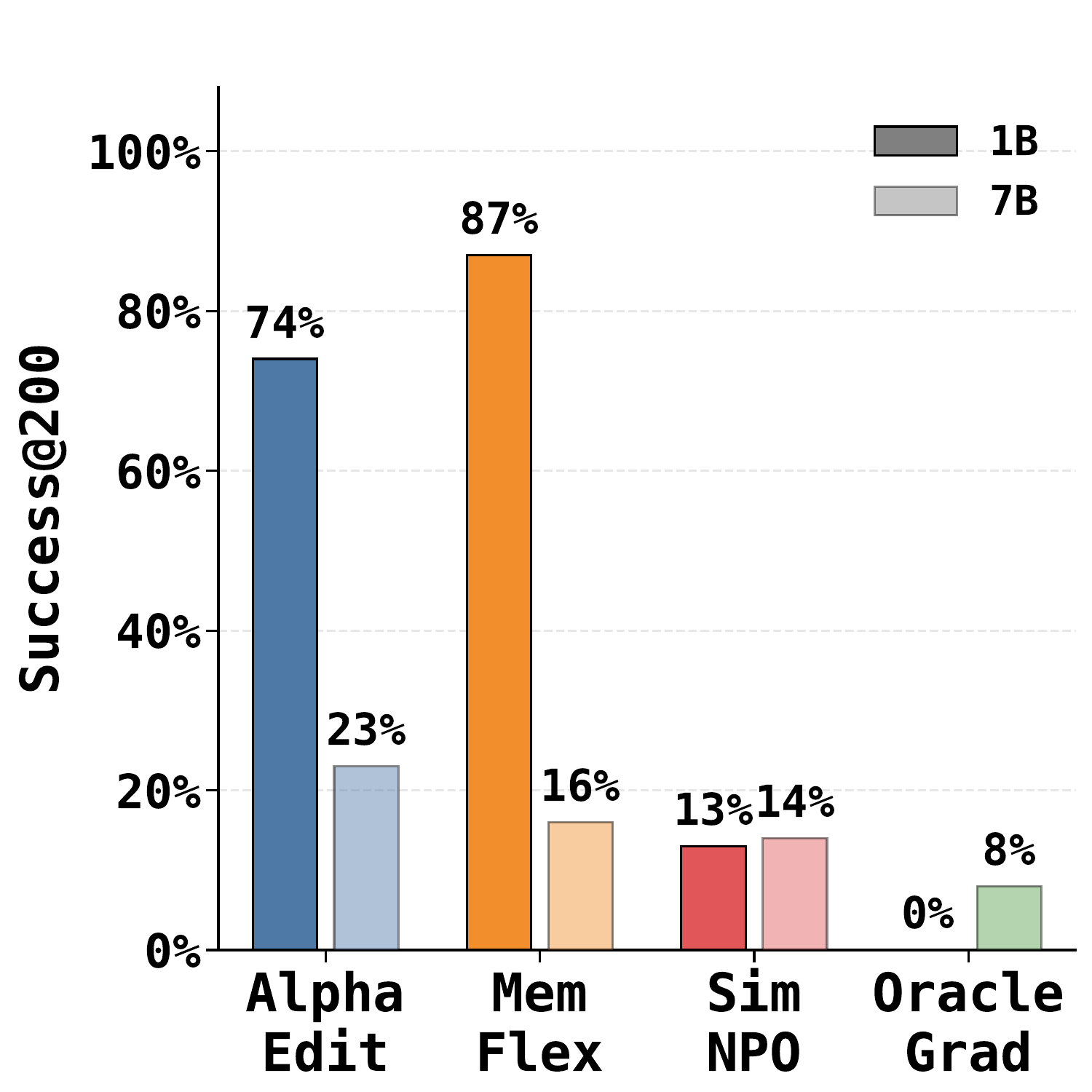}
    \caption{Relearning leakage rate.}
  \end{subfigure}
  \hfill
  \begin{subfigure}[t]{0.3\linewidth}
    \centering
    \includegraphics[width=\linewidth]{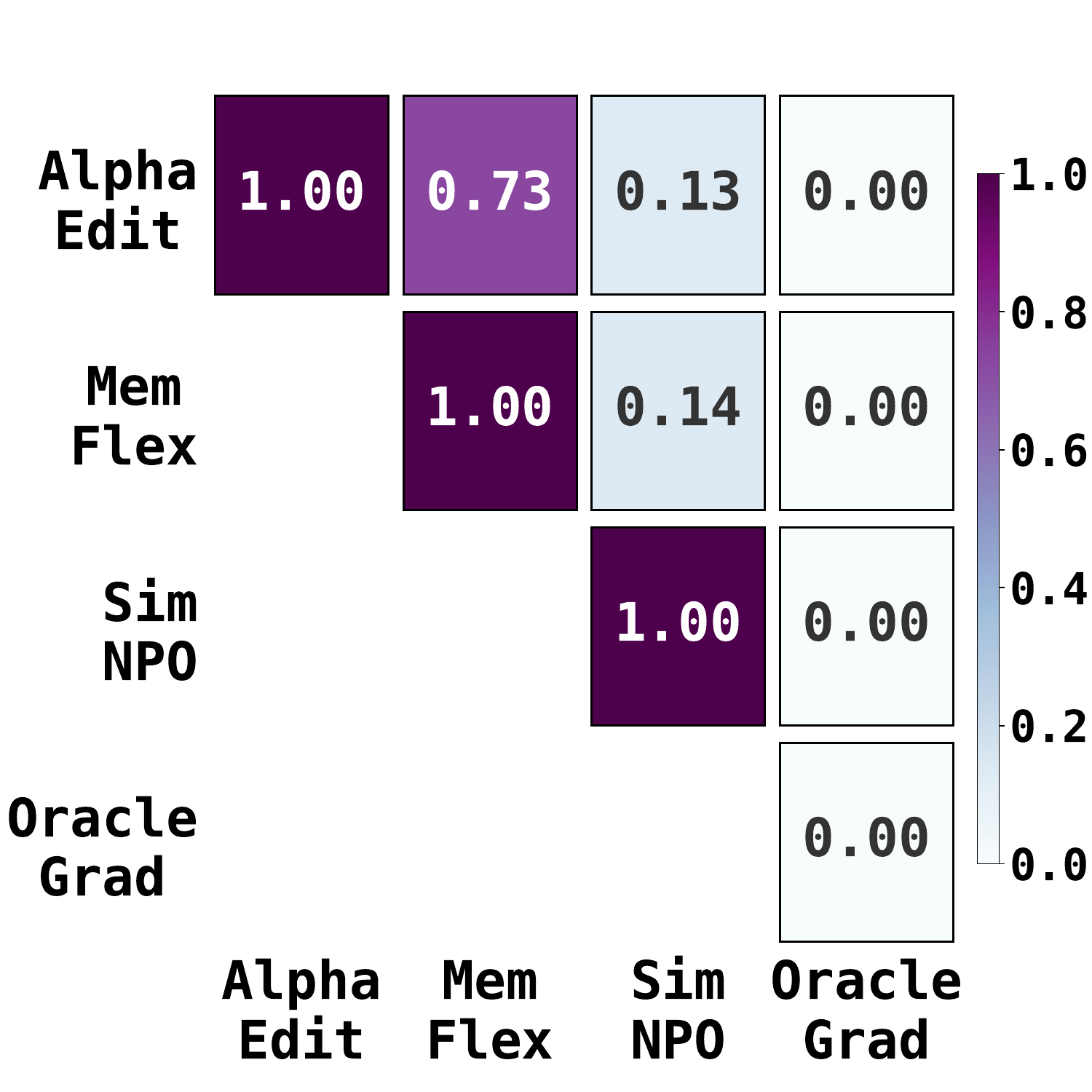}
    \caption{Leaked profiles Jaccard similarity - \olmosmall.}
  \end{subfigure}
  \hfill
  \begin{subfigure}[t]{0.3\linewidth}
    \centering
    \includegraphics[width=\linewidth]{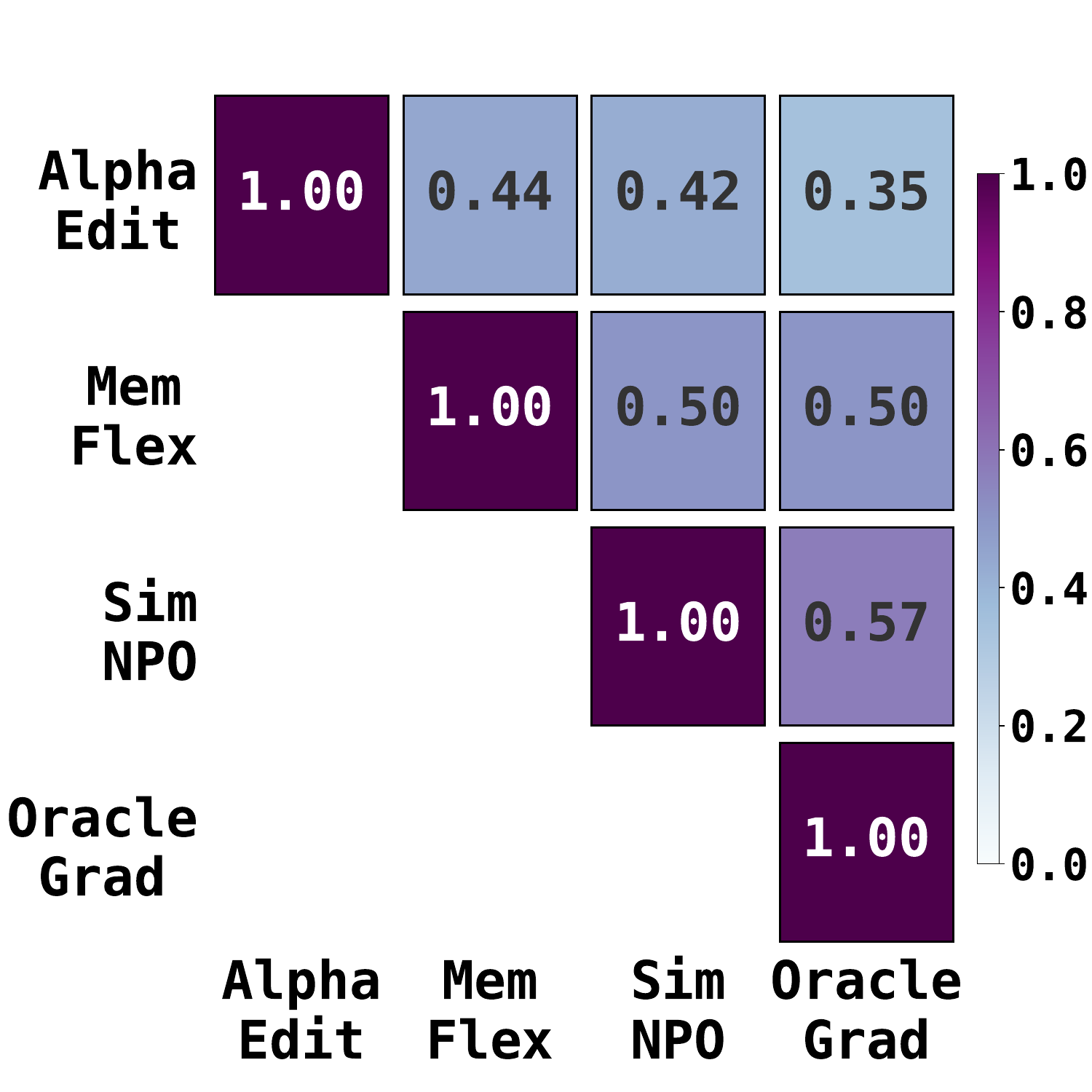}
    \caption{Leaked profiles Jaccard similarity - \olmobig.}
  \end{subfigure}
  \caption{Relearning vulnerability for \texttt{Email Address}.}
  \label{fig:relearn-email7b}
\end{figure}

\section{Hyperparameters - Tuning}
\label{appendix:tuning}

\subsection{Training \& Instruction Tuning}

Along with the discussion in \Cref{section:pii_injection}, we report here some additional details on the specific injection setup and instruction tuning.
\begin{table}[h]
\centering
\small
\caption{Training \& Instruction Tuning Hyperparameters}
\label{tab:training-hyperparams}
\begin{tabularx}{\linewidth}{>{\hsize=1.08\hsize}X>{\hsize=0.96\hsize}X>{\hsize=0.96\hsize}X}
\toprule
\textbf{Parameter} & \olmosmall & \olmobig \\
\midrule
Base model & \texttt{allenai/OLMo-2-0425-1B} & \texttt{allenai/OLMo-3-1025-7B} \\
Checkpoint revision & \texttt{step1907359-tokens4001B} & \texttt{step999000} \\
Instruction Tuning LoRA rank ($r$) & 16 & 16 \\
Instruction Tuning LoRA alpha ($\alpha$) & 8 & 8 \\
Instruction Tuning LoRA target layers & [14, 15] & [30, 31] \\
Instruction Tuning Early stopping & Best eval loss & Best eval loss \\
\bottomrule
\end{tabularx}
\end{table}

\subsection{Unlearning}
Unlearning hyperparameters were selected via grid search on a validation split (Driver's License forget / Email Address retain, cross-field), carried out independently for the \olmosmall and \olmobig models. The tuned hyperparameters for each method and model size are listed in Table~\ref{tab:tuning-unlearning}.

\paragraph{MemFlex.}
The primary tuning knobs were the forget/retain loss factors and the number of training epochs.

\paragraph{AlphaEdit.}
Extensive tuning was performed across edit magnitude (clamp norm, null-space threshold), optimization (gradient steps), covariance data source, target layers, and regularization. A key finding was that AlphaEdit cannot selectively target forget vs.\ retain data when both share the same QA format, as the activation subspaces overlap significantly.

\paragraph{SimNPO.}
The tuning for this method was relatively straightforward as it relies on more traditional gradient-based techniques.

\paragraph{OracleGrad.}
OracleGrad was tuned by comparing GradAscent (forget-only, no retain loss) against GradDiff (gradient ascent on forget + descent on retain). Gradient Difference (with the per-size forget/retain loss weights in Table~\ref{tab:tuning-unlearning}) provides high retain stability, while still allowing strong unlearning effectiveness.

\begin{table}[h]
\centering
\small
\caption{Unlearning -- tuned hyperparameters by method, selected independently for each model size. Values that differ between \olmosmall and \olmobig are in \textbf{bold}.}
\label{tab:tuning-unlearning}
\begin{tabularx}{\linewidth}{>{\hsize=0.71\hsize}X>{\hsize=1.40\hsize}X>{\hsize=\hsize}X>{\hsize=0.775\hsize}X}
\toprule
\textbf{Method} & \textbf{Hyperparameter} & \textbf{\olmosmall} & \textbf{\olmobig} \\
\midrule
\multirow{6}{*}{MemFlex}
  & Forget factor            & $-0.6$ & $-0.6$ \\
  & Retain factor            & $2.0$  & $2.0$ \\
  & Learning rate            & $\mathbf{3 \times 10^{-4}}$ & $\mathbf{1 \times 10^{-4}}$ \\
  & Gradient threshold       & $\mathbf{6 \times 10^{-4}}$ & $\mathbf{1 \times 10^{-5}}$ \\
  & Similarity threshold     & $0.92$ & $0.92$ \\
  & Epochs                   & $20$   & $20$ \\
\midrule
\multirow{7}{*}{SimNPO}
  & $\gamma$ (forget weight) & $3.0$  & $3.0$ \\
  & $\alpha$ (retain weight) & $\mathbf{0.01}$ & $\mathbf{0.5}$ \\
  & $\beta$ (sharpness)      & $10.0$ & $10.0$ \\
  & $\delta$ (margin)        & $1.5$  & $1.5$ \\
  & Learning rate            & $1 \times 10^{-4}$ & $1 \times 10^{-4}$ \\
  & Epochs                   & $200$  & $200$ \\
  & Retain loss / scheduler  & NLL / Constant & NLL / Constant \\
\midrule
\multirow{5}{*}{OracleGrad}
  & Method                   & GradDiff & GradDiff \\
  & $\gamma$ (forget weight) & $\mathbf{5.0}$ & $\mathbf{1.0}$ \\
  & $\alpha$ (retain weight) & $\mathbf{1.0}$ & $\mathbf{0.5}$ \\
  & Learning rate            & $\mathbf{1 \times 10^{-4}}$ & $\mathbf{5 \times 10^{-5}}$ \\
  & Epochs                   & $200$ & $200$ \\
\midrule
\multirow{11}{*}{AlphaEdit}
  & Clamp norm factor        & $0.5$ & $0.5$ \\
  & Null-space threshold     & $1 \times 10^{-3}$ & $1 \times 10^{-3}$ \\
  & $v$ gradient steps       & $50$ & $50$ \\
  & $v$ learning rate        & $5 \times 10^{-2}$ & $5 \times 10^{-2}$ \\
  & $v$ loss layer           & $\mathbf{15}$ & $\mathbf{31}$ \\
  & $v$ weight decay         & $\mathbf{1.0}$ & $\mathbf{0.5}$ \\
  & KL factor                & $\mathbf{1.0}$ & $\mathbf{0.0625}$ \\
  & L2 regularization        & $1.0$ & $1.0$ \\
  & MOM2 dataset             & \textbf{wikipedia + retain} & \textbf{wikipedia} \\
  & Target layers            & [4, 5, 6, 7, 8] & [4, 5, 6, 7, 8] \\
  & Batch size               & $\mathbf{5}$ & $\mathbf{1}$ \\
\bottomrule
\end{tabularx}
\end{table}

\end{document}